\documentclass[conference]{IEEEtran}
\IEEEoverridecommandlockouts
\usepackage{cite}
\usepackage{amsmath,amssymb,amsfonts}
\usepackage{algorithmic}
\usepackage{graphicx}
\usepackage{textcomp}
\usepackage{xcolor}

\usepackage{graphicx}
\usepackage{subfigure}
\usepackage{booktabs}
\usepackage{mathtools}
\usepackage{amsmath,amsfonts,amssymb,bm,amsthm}
\usepackage{multirow}
\usepackage{algorithm,algorithmic}
\newcommand\scalemath[2]{\scalebox{#1}{\mbox{\ensuremath{\displaystyle #2}}}}

\newtheorem{proposition}{Proposition}
\newtheorem{assumption}{Assumption}
\newtheorem{theorem}{Theorem}

\DeclarePairedDelimiter\floor{\lfloor}{\rfloor}
\DeclarePairedDelimiter\sround{\lceil}{\rfloor}
\newcommand{\showComments}{yes}
\newcommand{\note}[2]{
    \ifthenelse{\equal{\showComments}{yes}}{\textcolor{#1}{#2}}{} }

\newcommand{\sysname}{QSync}
\newcommand{\scenarioname}{hybrid-device training}
\newcommand{\concept}{quantization-minimized synchronous}

\newcommand{\figc}{Fig.~}

\def\BibTeX{{\rm B\kern-.05em{\sc i\kern-.025em b}\kern-.08em
    T\kern-.1667em\lower.7ex\hbox{E}\kern-.125emX}}
\pdfoutput=1
\begin{document}

\title{QSync:
Quantization-Minimized Synchronous Distributed Training Across Hybrid Devices}

\author{
\IEEEauthorblockN{Juntao Zhao\IEEEauthorrefmark{1}, 
                  Borui Wan\IEEEauthorrefmark{1}, 
                  Yanghua Peng\IEEEauthorrefmark{2}, 
                  Haibin Lin\IEEEauthorrefmark{2}, 
                  Yibo Zhu\IEEEauthorrefmark{2}, 
                  Chuan Wu\IEEEauthorrefmark{1}}
\IEEEauthorblockA{\IEEEauthorrefmark{1}The University of Hong Kong, Hong Kong}
\IEEEauthorblockA{\IEEEauthorrefmark{2}ByteDance Inc., USA}
\IEEEauthorblockA{juntaozh@connect.hku.hk}
}



\maketitle

\begin{abstract}
A number of production deep learning clusters have attempted to explore inference hardware for DNN training, at the off-peak serving hours with many inference GPUs idling. Conducting DNN training with a combination of heterogeneous training and inference GPUs, known as hybrid device training, presents considerable challenges due to disparities in compute capability and significant differences in memory capacity. We propose \sysname{}, a training system that enables efficient synchronous data-parallel DNN training over hybrid devices by strategically exploiting quantized operators. According to each device's available resource capacity, \sysname{} selects a quantization-minimized setting for operators in the distributed DNN training graph, minimizing model accuracy degradation but keeping the training efficiency brought by quantization. We carefully design a predictor with a bi-directional mixed-precision indicator to reflect the sensitivity of DNN layers on fixed-point and floating-point low-precision operators, a replayer with a neighborhood-aware cost mapper to accurately estimate the latency of distributed hybrid mixed-precision training, and then an allocator that efficiently synchronizes workers with minimized model accuracy degradation. \sysname{} bridges the computational graph on PyTorch to an optimized backend for quantization kernel performance and flexible support for various GPU architectures. Extensive experiments 
show that \sysname{}'s predictor can accurately simulate distributed mixed-precision training with $<5\%$ error, with a consistent $0.27-1.03\%$ accuracy improvement over the from-scratch training tasks compared to uniform precision.
\end{abstract}


\section{Introduction}

Production AI clouds typically include both training clusters and inference serving clusters: the former consists of GPU servers equipped with training GPUs (e.g., NVIDIA A100, V100) and the latter of servers with inference GPUs (e.g., NVIDIA T4, A10). The training cluster runs throughput-sensitive deep neural network (DNN) training jobs, while the inference clusters serve latency-intensive inference tasks with strict service level objectives (SLO).

Load on an inference serving cluster often exhibits strong daily patterns, with near-full-capacity consumption at daily peak hours and low usage valleys $(<40\%)$ at an off-peak time. On the other hand, training jobs in the training cluster often experience long queuing times. To expedite training jobs and improve the utilization of inference GPUs, \textbf{\scenarioname}, i.e., 
training using a mixture of training and inference GPUs has been proposed for exploiting unused resources in inference clusters to run training jobs.

Studies have addressed heterogeneous training, emphasizing either workload reallocation or elastic training methods. HetPipe~\cite{hetpipe} proposes a novel synchronized pipelined-parallel training approach to attain optimal workload balance. In contrast, AccPar~\cite{AccPar} concentrates on achieving equilibrium in the operator partition across devices in tensor-parallel training scenarios. Conversely, Aryl~\cite{aryl} implements a resource scheduling strategy that incorporates spare resources on inference GPUs to effectively execute training tasks.

However, the former approach heavily relies on the parallelism structure inherent to the training jobs and exhibits heightened sensitivity to communication bandwidth. Consequently, its suitability for data-parallel training jobs remains inadequate. 
In contrast, the latter approach is suitable for conventional heterogeneous device training scenarios. However, hybrid devices differ from normal case heterogeneous training with significant computation and memory discrepancies employed (refer to Sec.~\ref{sec:bg}). As a result, employing the same training setups (e.g. batch size) for both inference and training GPUs proves arduous when adapting to elastic methods.



\begin{figure}[t]
  \centering
  \includegraphics[width=\linewidth]{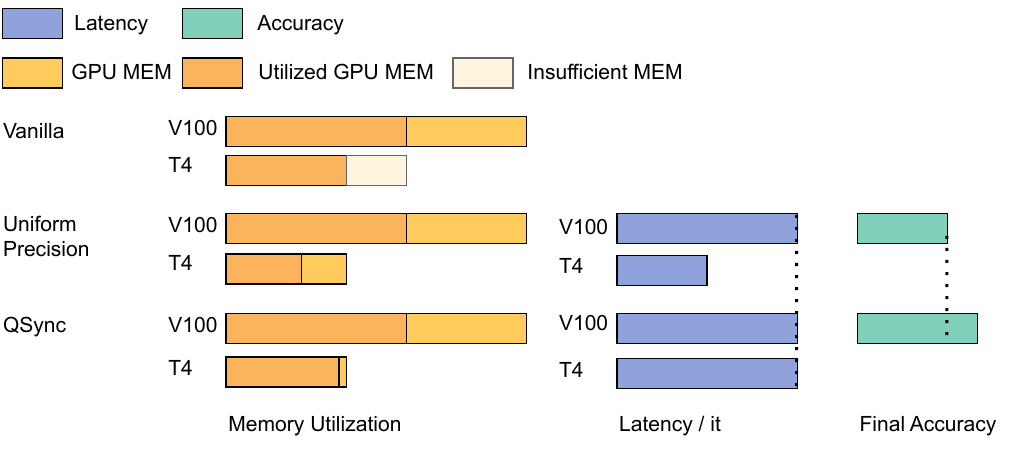}
  \vspace{-4mm}
  \caption{Illustration of \sysname{}. \sysname{} reduces the number of unnecessary quantized operators without sacrificing the overall training efficiency to recover model quality. } 
  \vspace{-3mm}
  \label{fig:methods_compare}
  \vspace{-3mm}
\end{figure}

Dynamic or variable batch sizing~\cite{Chen2020SemidynamicLB} 
is another approach to handling resource heterogeneity in data-parallel training tasks, which allocates a small local batch size to devices with smaller memory and computation capacities and a large batch size to the high-capacity ones, to balance the workload and training time. 
However, some operators (e.g. BatchNorm~\cite{BN}
) and training hyperparameter set-up rely on the local batch size
, e.g., the learning rate (lr) linearly scales with the batch size~\cite{lSRule}. Different batch size settings can significantly hurt the learned model quality (Sec.~\ref{sec:scenarios}). 

To circumvent the challenges associated with dynamic batch sizing, a promising approach entails the utilization of quantized operators. By employing low-precision computation and storage on inference GPUs, we can effectively mitigate memory requirements and minimize the disparity between inference and training GPUs, while simultaneously preserving the integrity of the local batch size to ensure accuracy. 
However, simply adopting a uniform low-precision (e.g., INT8) on inference GPUs may introduce much model accuracy degradation \cite{towards_mitigating}.  
A good trade-off between training efficiency and model accuracy should be carefully achieved by strategically selecting the precision for each operator on inference GPUs. 


We propose a \textit{\concept{}} training system, \sysname{}, that conducts effective \textit{hybrid mixed-precision training}, i.e., different GPUs hold different precision setups to train the same full precision model, over heterogeneous devices with minimal model accuracy degradation. 
As illustrated in \figc{}\ref{fig:methods_compare}, instead of using uniform low-precision for all computation operators in inference GPUs, \sysname{} intends to
\textbf{convert only necessary computation operators to their low-precision counterparts}. 
i.e., \sysname{} recovers the redundant low-precision operators in uniform low-precision quantization. The redundant low-precision operators refer to the operators that can be recovered to their higher bit-width representations, to improve the final model accuracy (Sec.~\ref{sec:pred}) while maintaining the global training throughput without introducing new overhead. 

The contributions of \sysname{} are summarized as follows.

$\triangleright$ We design a predictor that models the sensitivity of quantizable operators 
and accurately predicts the end-to-end latency of hybrid mixed-precision training. 
By applying stochastic quantization, we 
extend the previous theory to guarantee the convergence of the hybrid mixed-precision training, and give a proper 
model perturbation~\cite{hawq-v3} indicator on different low-precision operators based on it. 
Through profiling, the predictor carefully models the casting cost (cost of converting tensors between different precisions) and tackles neighboring dependent cascading precision change for operators. Experiments show that \sysname{}'s predictor gives an indicator of precision selection that outperforms the existing schemes and can accurately simulate hybrid mixed-precision training with $<5$\% average error in terms of throughput prediction.

$\triangleright$ We design an efficient allocator to assign precisions to different operators on heterogeneous devices, achieving quantization-minimized distributed synchronous training. 
The allocator searches operators' precision settings starting from the fastest available precision setup 
that minimizes the local model execution latency under the device memory constraints. Based on the perturbation indicator, the allocator then recovers part of the operators' precision with a higher bit. Experiments show that 
with the allocator, unnecessary low-precision operators on inference GPUs can be recovered, with up to $27$\% overall training efficiency gain compared to dynamic batch sizing 
and up to 1.03 \% model accuracy improvement to the uniform low-precision scheme.



$\triangleright$ 
We bridge the computation graph of \sysname{}'s from PyTorch to our own customized backend, named \textit{LP-PyTorch}, which supports and promotes data type versatility for CUDA training. LP-PyTorch provides templated and tunable access to the underneath training kernels (e.g., CUTLASS / CUDNN \cite{cutlass}). It supports most of the existing GPU structures and data versatility. It further optimizes the pipelining of fixed-point kernel execution and achieves $>10\%$ end-to-end performance gain for the INT8 training. The code is available at https://github.com/bytedance/QSync.git.



\section{Background and Motivation}\label{sec:bg}

\subsection{Hybrid-device training}
\label{sec:scenarios}
In production AI systems, training clusters, which run resource-intensive DNN model training jobs, are typically heavily loaded at all times. Inference serving clusters, which serve latency-sensitive online model queries, 
commonly exhibit daily usage patterns according to peak/off-peak hours of applications driven by the DNN models~\cite{aryl}.

\begin{figure}[t]
  \centering
  \includegraphics[width=\linewidth]{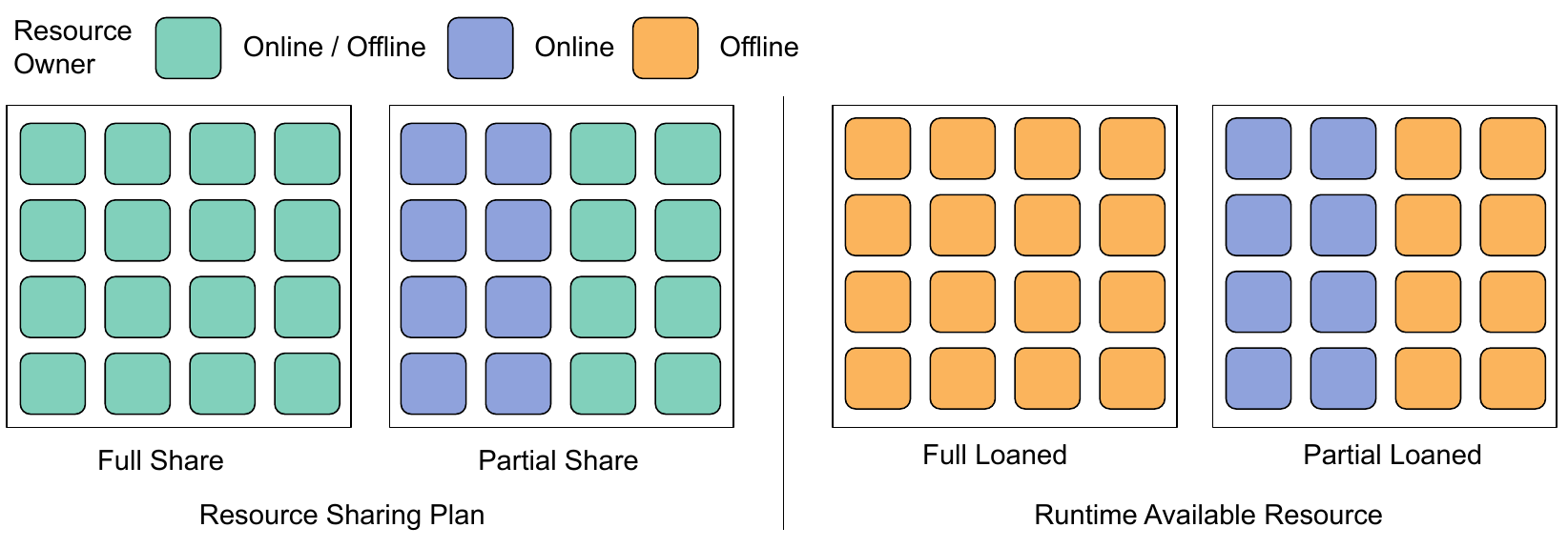}
  \vspace{-5mm}
  \caption{Full and partial resource sharing. \textbf{Left}: Full-sharing GPU has no strict resource isolation but the partial share has a strict resource reservation. \textbf{Right}: In training, the resource on the full-sharing inference GPU can be fully utilized for the training job. As opposed to this, in partial resource sharing, only a portion of the resource is made available.} 
  \label{fig:gpu_sharing}
  \vspace{-5mm}
\end{figure}

\textbf{Hybrid training} can be considered as a distinctive instance of heterogeneous training, where both training resources and underutilized resources from inference clusters are leveraged for executing DNN training tasks. 
To share GPU resources in inference clusters, 
\textbf{full-sharing} provides the whole GPU for training jobs, 
while \textbf{partial-sharing} preserves some GPU resources (e.g., memory, GPU threads) for online inference serving with the rest for training jobs. 
An illustration is given in \figc{}\ref{fig:gpu_sharing}. Isolation of resources in the partial-sharing mode is typically achieved through Multi-Process Service (MPS)~\cite{pipeswitch}. Typically, the low-caliber inference GPU has much lower memory and compute capability compared with the training GPU, and partial isolation makes the situation much worse. 


Existing research advocates for heterogeneous training, proposing approaches such as pipelined parallelism (PP) synchronization~\cite{hetpipe} and tensor parallelism (TP) for partitioning tensors among heterogeneous devices~\cite{AccPar}. However, it is worth noting that these approaches heavily rely on specific parallelism structures, such as parameter server and pipelining for HetPipe, and Tensor Parallel for the AccPar. Consequently, adapting these methodologies to alternative parallelism structures, such as data parallelism, necessitates substantial effort and modifications. Furthermore, while TP and PP demonstrate commendable performance for large-scale models, they tend to impose higher communication requirements and exhibit heightened sensitivity to bandwidth limitations, particularly when devices are distributed across different clusters.

Other studies, exemplified by Aryl~\cite{aryl}, approach the training workload as individual jobs and focus on scheduling these workloads across a combination of training and inference devices. For instance, they achieve this by scaling the number of workers while maintaining a fixed local batch size across the devices. However, it is imperative to note that in the context of \scenarioname, the training and inference GPUs (e.g., NVIDIA V100 vs. T4) exhibit significant disparities in terms of memory capabilities, as evidenced in Table~\ref{table:tops_gpus}. Furthermore, this discrepancy is exacerbated by the sharing configuration depicted in Fig.~\ref{fig:gpu_sharing}. A training setup, such as a specific batch size, which is compatible with a training GPU, may not directly translate to an inference GPU. The mismatch can result in memory overflow or give rise to substantial synchronization bubbles, thus squandering the resources of the training GPU.

Dynamic batch sizing~\cite{Chen2020SemidynamicLB} handles a heterogeneous training environment by adjusting the batch sizes according to device capacities. While maintaining a constant global batch size, devices with higher capacities handle larger local batch sizes while low-capacity devices process data of smaller batch sizes, 
 to achieve load-balancing. However, a key issue is not addressed in the existing dynamic batch sizing designs, changing the batch size may influence the training semantics such as convergence efficiency and final model accuracy. 
 For example, batch normalization (BN) collects and updates statistical information within a batch, and its results depend heavily on how the data is grouped into batches; 
 the hyperparameter settings (e.g., momentum $\lambda$) 
  and the statistical result of the moving average in BN 
 (running mean and running variance) highly depends on the batch size~\cite{Wu2021RethinkingI}. To address this problem, some works use sync-bn~\cite{sync_bn}, 
 which forces a synchronization among the statistical result above, but introduces additional synchronization overhead. 
 As a result, expertise is required to adjust the original setting, 
 such as a starting learning rate, a learning rate scheduler, and even incorporating an additional model structure. Our experiment in Sec.\ref{sec:eval} demonstrates that when using the learning rate adaptation setting proposed by existing works\cite{Chen2020SemidynamicLB}, dynamic batching still results in significant degradation for from-scratch training but also decreases the overall training throughput.  



\begin{table}[t]
    \centering
    \caption{Capability of Different Devices}
    \label{table:tops_gpus}
    \resizebox{\linewidth}{!}{%
    \begin{tabular}{c|ccc|c}
    \toprule 
        GPU & FP32 TFLOPS & FP16 TFLOPS & INT8 TOPS & Memory \\ 
    \midrule
        T4 & 8.1 & 65 & 130 & 16G\\ \hline
        V100 & 15.7 & 125 & / & 32G \\  
    \bottomrule
    \end{tabular}%
    }
    \vspace{-5mm}
\end{table}

\vspace{1mm}
\noindent \textbf{Opportunity: Using mixed-precision operators for \scenarioname{} without changing the batch size}. 

To keep the batch size settings unchanged, an alternative way is compression. In particular, quantized distributed training (QDT) is widely studied. Quantization compresses model weights and activations by mapping high-precision values to low-precision equivalents, which saves the memory required by the model weight and activation but also speeds up the training process. Table~\ref{table:tops_gpus} shows the tera (floating point) operations per second (TOPS / TFLOPS) of operations at different precisions on NVIDIA T4 and V100 GPUs~\cite{V100, T4}. The TOPS increases when the precision is halved, exhibiting substantial computation acceleration by using low-precision operators. This is also true on other chips such as NVIDIA A10 and A100. In the realm of DNN training, FP16/BF16 automated mixed-precision training has found widespread application across various tasks. Furthermore, QDT has made remarkable progress by pushing the boundaries to include int8 quantized distributed training~\cite{Markov2023QuantizedDT}. While existing distributed quantized training methods have successfully reduced memory requirements and expedited the training process, the straightforward uniform quantization of weights can lead to compromised theoretical convergence. Consequently, this compromises the accuracy and convergence rate~\cite{towards_mitigating}.

Existing quantized distributed training methods uniformly quantize all main operators (linear, conv) to the same precision across different devices, which is insufficient when facing \textit{\scenarioname}. Firstly, we typically share a training job with a batch size that conforms to the training GPU. This ensures that the training GPU always has enough memory to hold the batch and eliminates the need to quantize the operators on it, also, some lower-precision (e.g. INT8) may be not supported by the training hardware (e.g. V100), as shown in Table~\ref{table:tops_gpus}. Secondly, we only need to perform the necessary quantization on the inference GPUs to keep the training efficiency while minimizing accuracy degradation.

Our objective is to leverage quantized operators to achieve memory reduction and accelerate the training process while placing a strong emphasis on preserving accuracy. Rather than applying uniform quantization to all operators, our approach focuses on selectively quantizing essential operators specifically on inference GPUs. We quantize enough operators to fit the training workload into the inference GPU but leave some operators unchanged or with higher precision to mitigate the speed differential between inference and training GPUs after quantization and improve accuracy. We term this approach as \textit{\concept}. Our particular focus lies in data-parallel training jobs configured on training GPUs, particularly the batch size, and strive to execute them within a hybrid device environment. This endeavor presents novel challenges and opens up new opportunities for design exploration.



\subsection{Challenges in \concept{} in \scenarioname{}}



\noindent
\textbf{Measurement of quantization impact on model accuracy and training efficiency.} 
To simultaneously consider model accuracy and training throughput induced by low-precision operators in \scenarioname{}, we need to know how changes in operator precision impact model accuracy and training throughput. 
The accuracy impact of both floating-point and fixed-point low-precision operators must be taken into account. Previous studies~\cite{fixed_back, hawq-v3} only address one of them and focus on the forward pass in DNN training, while 
operator precision in backward propagation should also be considered. For throughput estimation, accurate modeling and prediction of the timeline view of the mixed-precision global training are needed. 
Campo~\cite{campo} used performance modeling to predict the casting cost and operational performance with low precision and introduced a cost-aware graph rewriting strategy to optimize mixed precision training. 
It only considers casting costs between FP32 and FP16, 
ignores the precision dependency between operators\footnote{There exists CUDA Ops that promote the widest input type. For example, the precision of the {\em add} operator depends on the largest precisions of its inputs. If the precisions of two inputs are not the same (such as FP16 and FP32), a cast operator is added to convert the lower-precision input to the higher precision for addition. Other operators like {\em ReLU} that are not directly handled by the auto casting also depend on its input precision as well.} and cannot reflect the overall model runtime in the distributed setting. 


\vspace{1mm}
\noindent \textbf{Efficient precision allocation.} Given possible precisions and 
a large number of precision-adjustable operators in a DNN model, it is time-consuming 
to brute-forcibly search for the best setting over all feasible mixed-precision settings. 
For example, given INT8, FP16, and FP32 as three optional precisions and considering setting precisions for 73 linear operators in BERT or 52 Conv2D operators in a ResNet50, the search space is $3^{52}$ or $3^{73}$, respectively. Efficiently and correctly derive the optimized mixed-precision settings is challenging. 



\vspace{1mm}
\noindent \textbf{Low-precision versatility supports}. The realization of the aforementioned advantages of quantized training relies on the effectiveness of low-precision operator kernels. The existing training framework (e.g. PyTorch, Tensorflow) usually supports FP16/FP32 CUDA training by default. To bridge operators to their extended low-precision implementations, 
an efficient pipeline is required to access low-precision kernels and tune their performance for different hardware. For inference serving scenarios, \textit{kernel tempting} is discussed to tune operators for different target devices 
\cite{Xu_AITemplate_2022}. On the contrary, none of the existing frameworks support templating low-precision training kernels (e.g. backward ops), let alone adapting them to hardware for optimized performance.

\section{Overview}

\begin{figure}[t]
  \centering
  \includegraphics[width=\linewidth]{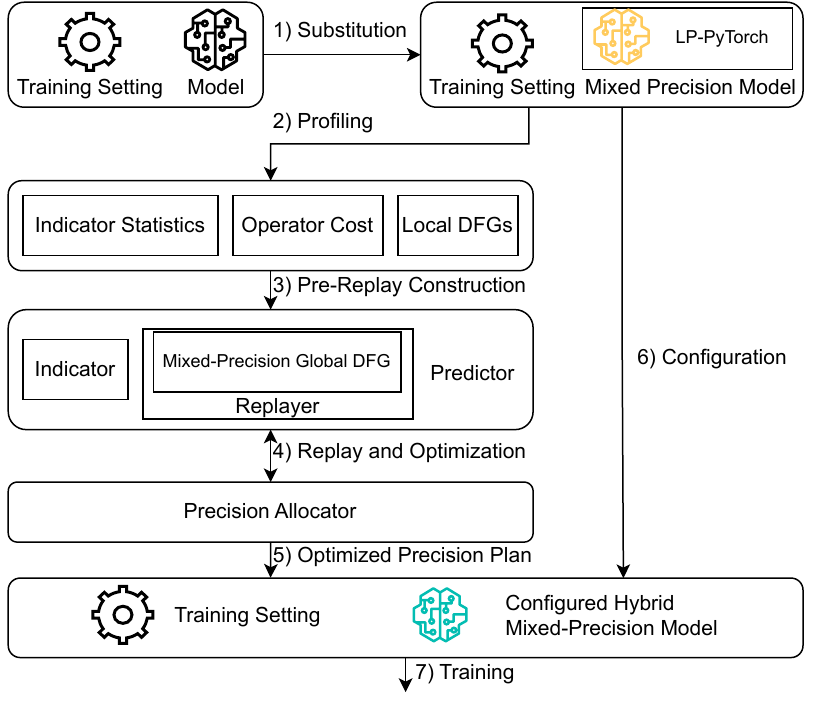}
  \caption{QSync Workflow}
  \label{fig:architecture}
\end{figure}

We propose \sysname{}, a \concept{} distributed training system to enable efficient synchronous data-parallel DNN training over hybrid devices. We consider a training cluster equipped with training GPUs of the same type and an inference serving cluster with inference GPUs of the same type. A distributed DNN training job can leverage multiple training GPUs and available resources (e.g., memory, compute capability
) on some inference GPUs. Fig.~\ref{fig:gpu_sharing} gives an overview of the workflow of \sysname{}. 
The main idea of \sysname{} is to use low-precision formats for necessary operators on inference GPUs. The workflow of \sysname{} goes as follows. 1) Substituting operators in a model with mixed-precision implementation for the target hardware. 2) The cost and memory requirements for the operators under different precision are collected through profiling. Statistical data, like model depth, tensor dimension, and norms are also collected by running a few iteration steps on the GPUs using smaller batch sizes. Local DFGs with communication dependency are also traced by constructing homogenous GPU sub-sets. 
3) The predictor calculates the indicator result (model perturbation) based on statistical data of the operators and builds a global mixed-precision data flow graph. 
4) The precision allocator, with the help of the Replayer in Predictor, simulates distributed model training and estimates overall throughput under different mixed-precision settings using the guidance of the Indicator, and greedily optimizes the precision allocation plan starting from the initial global DFG.
5) The optimized precision plan is then fed back to the mixed-precision training system. 6) The mixed-precision backend then configures the low-precision kernel by selecting the best device-optimized configuration 
7) Hybrid mixed-precision distributed model training is carried out using the optimized precisions. 
\noindent
\sysname{} includes three main modules: 

\noindent\textbf{The Predictor} 
is composed of an Indicator that generates operator perturbation towards precision, and a Replayer simulates local and distributed training and estimates the memory consumption and per-iteration training time. 

\vspace{1mm}
\noindent\textbf{Precision Allocator} interacts with the predictor to search for better precision settings for operators on the inference GPUs. 


\vspace{1mm}
\noindent\textbf{LP-PyTorch} is a backend that enhances the capabilities of deep learning frameworks like PyTorch to efficiently tune and run low-precision kernels. 




\section{The Predictor}\label{sec:pred}
The main problem addressed by \sysname{} is to find an optimized precision allocation plan to operators on inference GPUs that minimizes the model accuracy 
degradation introduced by low-precision kernel execution while 
maintaining the training throughput. This formulates the problem below:
\vspace{-2mm}
\begin{equation}\label{eq:obj-funct}
\vspace{-2mm}
\scalemath{1}{
\begin{aligned}
\min_{\{b_{io}\,\mid\,{i \in \mathcal{K}_{inf}}, {o \in O} \}} \quad & \sum^{}_{i \in \mathcal{K}_{inf}} \sum^{}_{o\in O} \Omega_{o}^{(b_{io})}\\
\textrm{s.t.} \quad &  \mathcal{M}_i(\{b_{io}\,\mid\, o \in O \}) \leq M_{i}^{max}, \forall i \in \mathcal{K}_{inf} \\
\quad & \mathcal{E}(\{b_{ko}\,\mid\, k \in \mathcal{K}, o \in O \}) \geq T^{min}   
\end{aligned}
}
\end{equation}

\noindent In our \scenarioname{} among $\mathcal{K}$ GPUs, $\mathcal{K}_{inf}$ is the set of inference GPUs. $\Omega_{o}^{(b_{io})}$ is operator $o$'s sensitivity with bit precision $b_{io}$ on inference GPU $i$, and $O$ is the set of all operators in the model directed acyclic graph (DAG). 
$M_{i}^{max}$ is the available memory capacity on inference GPU $i$. $T^{min}$ is the 
training throughput of the DNN training job that can be obtained using the same low-precision for all operators on inference GPUs under the memory constraints, e.g., converting all operators to int8 or fp16 depending on the lowest precision that the inference GPUs support. 
$\mathcal{M}_i(\cdot)$ is the predictor function that estimates the memory consumption by the training job on inference GPU $i$. $\mathcal{E}(\cdot)$ estimates overall training throughput based on the precision plan $\{b_{ko}\,\mid\, k \in \mathcal{K}, o \in O \}$ among all GPUs. Especially, $b_{ko}=32$ on each training GPU k $\in \mathcal{K} \setminus \mathcal{K}_{inf}$. Solving problem~(\ref{eq:obj-funct}) poses new challenges: 1) How to build an effective sensitivity indicator $\Omega_{o}^{(b_{io})}$ to measure the relationship between model accuracy degradation and operator precisions. Low-precision operators affect both forward and backward passes and can be fixed-point or floating-point, which complicates the theoretical analysis. 2) How to construct accurate predictors $\mathcal{M}_i(\cdot)$ and $\mathcal{E}(\cdot)$. Due to the casting cost between different precision 
and precision-dependent operators whose execution latency and memory cost depend on their inputs precision, 
together with the presence of a communication operator and its dependency, the end-to-end model training latency cannot be readily expressed as an independent summation of sequential operator execution costs. 

When the precision of the forward operation is changed, the execution of the corresponding backward operation is also changed due to the casting. This means that precision change can lead to modifications in both the forward and backward passes of a given operator. For this reason, in this paper, we refer to an \textbf{operator} as a pair of forward and backward operations, and \sysname{} alters the precision of forward and backward operations together.

\subsection{Indicator}\label{sec:Ind}
The perturbation indicator $\Omega_{o}^{(b_{io})}$ qualifies the relationship between model accuracy and operator precisions. 
We explicitly examine the variance of the gradient of model weights introduced by the low-precision casting of operators. Large gradient variance can lead to large and unstable weight updates, making training difficult to converge. 
Following ACTNN~\cite{chen2021actnn}, we analyze the convergence of our \scenarioname{} when using an Unbiased Quantizer (UQ). 
UQ quantizes an original number to its unbiased estimation, i.e., $\mathop{\mathbb{E}}[UQ(x)] = x$. Stochastic rounding (SR) performs non-deterministic rounding according to the residual to nearby integer values, which is unbiased. 
We apply SR $\sround{\cdot}$ as our rounding method for quantization. $f(\cdot)$ denotes the loss function in model training with learning rate $\eta$. Consider the empirical-risk minimization of the loss function with parameter $\mathbf{x} \in R^{d}$ on a training dataset $D$. The DNN training problem can be modeled as:
\begin{equation}\label{eq:expectation_diff} 
    \min_{\mathbf{x} \in R^{d}} f(\mathbf{x}) \coloneqq \mathop{\mathbb{E}}_{s \sim D}[f_s(\mathbf{x};\{b_{io} \,\mid\, o \in O \})], \forall i \in \mathcal{K}
\end{equation}

\noindent where $s$ is a random sample from dataset $D$ and $f_s(\mathbf{x};\cdot)$ is the local loss function with a precision set-up for weight $\mathbf{x}$ on the GPU that processes the sample. $f_s^{(0)}(\mathbf{x})$ denotes a local loss function without low-precision operators. In \sysname{}, we consider representative loss functions (combination) such
as mean square error (MSE) and cross-entropy (CE) with softmax. Denote the input to these loss functions to be $v^{(L)}$, the ground truth to be $y^{(L)}$, the corresponding gradient function of input $v^{(L)}$ respect to loss function can be expressed as: $\nabla{v^{(L)}} = \gamma  (v^{(L)} - y^{(L)})$ where $\gamma \in \{\frac{2}{N}, \frac{1}{N}, -1\}$. 
In \sysname{}, the precision of these loss functions in the DNN model graph is unchanged. 
The following proposition states that we can obtain unbiased gradient estimation under the specification of \sysname{}: 
\begin{proposition}[Unbiased Gradient] 
With the loss function unchanged, by using an unbiased quantizer for linear operators 
, we have $\mathop{\mathbb{E}}[\nabla f_s(\mathbf{x};\{b_{io}\,\mid\, o \in O \}))] = \mathop{\mathbb{E}}[\nabla f_s^{(0)}(\mathbf{x})]$.
\end{proposition}

\noindent Assuming SGD training, convergence proof of \sysname{}'s mixed-precision training can follow ACTNN~\cite{chen2021actnn}. 
Consider an initial model parameter $\mathbf{x}_{0}$ with several convergence assumptions that are widely used~\cite{basu2019qsparse, dont_waste_bit}.

\begin{assumption}\label{assum:conver}
\small
$\forall \mathbf{x}_t, \mathbf{x'}_t \in R^d$ in the $t$-th training iteration:

\noindent\textbf{A.1} ($\mathcal L_2-Lipschitz$) $||\nabla f(\mathbf{x}_t)-\nabla f(\mathbf{x'}_t)|| \leq L||\mathbf{x}_t-\mathbf{x'}_t||$;

\noindent\textbf{A.2} (\emph{existence of global minimum}) $\exists\ f^{\ast}\ s.t.\ f(\mathbf{x}_t) \ge f^{\ast}$;


\noindent\textbf{A.3} (\emph{bounded variance}) There exists $\sigma^2 > 0$, s.t.$Var[\nabla{\mathbf{x}}] \leq \sigma^2 \forall \mathbf{x}$
\end{assumption}

\begin{theorem}\label{th:convergence}[Convergence] 
Let $T$ be the maximum number of iterations. Under Assumption~\ref{assum:conver}, we have
\vspace{-3mm}
\begin{equation*}
\scalemath{0.8}{
\begin{aligned}
     \min_{t=0,1,\ldots,T-1}\mathop{\mathbb{E}}[\|\nabla{f}(\mathbf{x}_t)\|^2] &\leq \frac{f(\mathbf{x}_{0}) - \mathop{\mathbb{E}}[f^{*}]}{\sum_{t=0}^{T-1}(-\eta + \eta^2 \frac{L}{2})} + \frac{\sum_{t=0}^{T-1}\eta^2 \frac{L}{2}\sigma^2}{\sum_{t=0}^{T-1}(-\eta + \eta^2 \frac{L}{2})} 
\end{aligned}
}
\end{equation*}
\end{theorem}

\noindent 
Except for $\sigma$, theorem~\ref{th:convergence} has the same form as FP32 that ensures the convergence and training ability of \sysname{}. $\sigma$ shapes the converged solution.

For a scalar $x$, with fixed-point quantization, $\bar{x} = \frac{x - z_x}{q_x}$ and $\hat{x} = \sround{\bar{x}} \times q_x + z_x$, where $\bar{x}$ is a contiguous number obtained by scaling $x$ with zero-point $z_x$ and scaling factor $q_x$, $\sround{\bar{x}}$ is the quantized scalar and $\hat{x}$ is the contiguous dequantized scalar. For floating-point quantization, \cite{fp8training}, the value of a scalar is represented by $x = s \cdot 2^e \cdot (1 + m )$, where $s, e, m$ are a sign, effective exponential bit and the mantissa. The exponents' bits are truncated and stochastic rounding is applied to the mantissa. For any vector $\mathbf{x}$, it has the following characteristics: 

\begin{proposition}[Tensor Quantization Variance] $Var[\hat{\mathbf{x}}] = \frac{q_{\mathbf{x}}^2D_{\mathbf{x}}}{6}$ for fixed-point quantization. $Var[\hat{\mathbf{x}}] = \frac{2^{2e} \epsilon^2 D_{\mathbf{x}}}{6}$ for floating-point quantization. $D_{\mathbf{x}}$ is the dimensionality of tensor $\mathbf{x}$. 
\label{prop:tensor_quant_var}
\end{proposition}

\noindent $\epsilon$ here is $2^{-k}$. For the IEEE standard floating-point formats, $k = 9$ for float16. We next model the variance increment of different operators in a 
DNN model, considering parameter, activation, and gradient in the forward and backward passes. 

\begin{proposition}[Variance Increment] We have the variance increment of operator ${o}$ with bit precision $b_o$, $\Omega_{o}^{(b_o)}$, as
\begin{equation}
    \scalemath{0.8}{
    \Omega_{o}^{(b_o)} = \gamma^2 d_o\hat{\sigma}_{fp}^{(o)} + (d_L-d_o) \hat{\sigma}_{bp}^{(o)}
    }
\end{equation}

\noindent Especially, for unary-input computation-intensive operators (e.g., Linear, Convolution) 
, we have
\vspace{-1mm}
\begin{equation}\label{eq:fp_ind}
\scalemath{0.8}{
\hat{\sigma}_{fp} = \left\{
\begin{aligned}
&\frac{1}{6} (\|{\mathbf{x}}\|^2 q_{\tilde{{\mathbf{v}}}}^2D_{{\mathbf{v}}} + \|\hat{{\mathbf{v}}}\|^2 q_{{\mathbf{x}}}^2D_{{\mathbf{x}}}), &\ \mbox{\small fixed-point quantization}, \\
& \frac{1}{6}\epsilon^2(\|\mathbf{x}\|^22^{2e_{\mathbf{v}}}D_{\mathbf{v}} + \|\hat{{\mathbf{v}}}\|^22^{2e_{\mathbf{x}}}D_{\mathbf{x}}), &\ \mbox{\small floating-point quantization}
\end{aligned}
\right.
}
\end{equation}

\vspace{-7mm}

\begin{equation}\label{eq:fixed_ind}
\scalemath{0.8}{
\hat{\sigma}_{bp} = \left\{
\begin{aligned}
&\frac{1}{6}(\|\nabla{{\mathbf{v}}}\|^2q_{\tilde{{\mathbf{v}}}}^2D_{{\mathbf{v}}} + \|\hat{{\mathbf{v}}}\|^22^{2e_\nabla{_{\mathbf{v}}}}\epsilon^2D_\nabla{_{\mathbf{v}}}), &\ \mbox{\small fixed-point quantization}, \\
& \frac{1}{6}\epsilon^2(\|\nabla{\hat{{\mathbf{v}}}}\|^22^{2e_{\mathbf{v}}}D_{{\mathbf{v}}} + \|\hat{{\mathbf{v}}}\|^22^{2e_\nabla{_{\mathbf{v}}}}D_\nabla{_{\mathbf{v}}}), &\ \mbox{\small floating-point quantization}.
\end{aligned}
\right.}
\end{equation}
\end{proposition}

\noindent Here ${\mathbf{v}}$ and ${\mathbf{\nabla{v}}}$ are the activation and gradient of activation for the operator $o$. 
From the observation of equation~\ref{eq:fp_ind} and~\ref{eq:fixed_ind}
, an operator's sensitivity to the precision is determined by its depth $d_o$ with respect to the model depth $d_L$. The depth of an operator inside a model forward DAG is a measure of its distance from the root node, which can be directly obtained by applying depth-first search
. Tensor dimensionality $D_{{\mathbf{v}}}, D_\nabla{{\mathbf{v}}}$ and $D_{\mathbf{x}}$, and norms $\|\nabla{{\mathbf{v}}}\|^2, \|\nabla{\hat{{\mathbf{v}}}}\|^2, \|\hat{{\mathbf{v}}}\|^2$ also matters. For fixed-point operators, the scaling factors 
of the quantization affected input and weight, $q_{\tilde{{\mathbf{v}}}}, q_{{\mathbf{x}}}$
, also contribute to the variance. $e_{\mathbf{v}}$, $e_{\mathbf{x}}$, $e_\nabla{{\mathbf{v}}}$ are effective bits, which can be derived with the data's magnitude (maximum and minimum). 
 These factors 
 can be collected through profiling. 
The norms and scaling factors $q_{\tilde{{\mathbf{v}}}}$ and $q_{{\mathbf{x}}}$ are changing during training, so $\Omega_{o}^{(b_o)}$ of the operator is changing. 
Our experiments (Sec.~\ref{sec:indicator_trace}) show that most of the relative values of factors related to the training process do not change significantly. To improve efficiency, we use the running mean of the first 50 iterations as the perturbation result of model operators, we also half the training batch size for profiling these results.
Unary operators with one argument 
such as MaxPool does not hold learnable parameters. Their variance is only introduced in the forward pass and bounded by their input's tensor quantization variance bound. Their $\hat{\sigma}_{bp}^{(o)}$ is zero. Specifically, \sysname{} does not modify pure matmul operations, which involve binary inputs. 

\subsection{Replayer}

\begin{figure}[t]
  \centering
  \includegraphics[width=\linewidth]{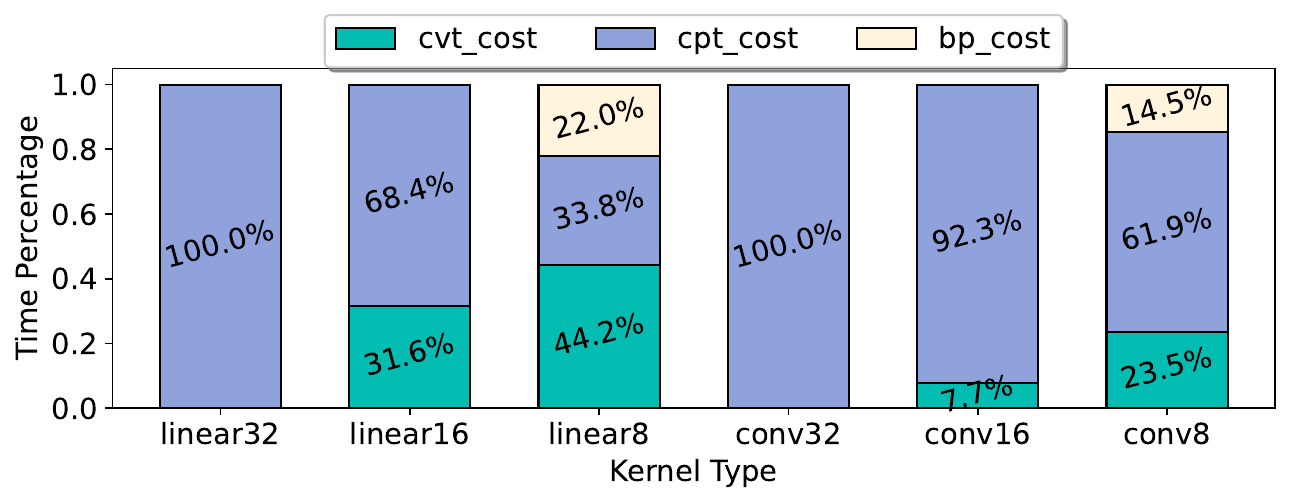}
  \vspace{-5mm}
  \caption{Cost Composition of an Operator 
  }
  \vspace{-4mm}
  \label{fig:cost_percent}
\end{figure}

\noindent\textbf{Cost Mapper.} 
To obtain the training throughput estimation $\mathcal{E}(\cdot)$ under different mixed-precision settings, we  
first analyze the cost composition of an operator at different precisions in each training iteration. Take the second last convolution in VGG16 and a regular linear operator from one of the attention blocks in BERT as examples: we execute the operator 100 times on T4 
and obtain the average time composition as shown in Fig.~\ref{fig:cost_percent} for $INT8,FP16\ and\ FP32$. 
$cvt\_cost$ denotes the casting overhead in the forward pass in converting input and weight tensors to low precision, 
$bp\_cost$ is the additional casting overhead\footnote{Integer backward computation is shown to incur low efficiency \cite{cutlass}. In \sysname{}, we perform the backward computation of fixed-point kernels in FP16, which incurs additional casting costs.} in backward computation, and $cpt\_cost$ is the operator's execution time including both forward and backward computation. The casting cost is non-negligible with low-precision operators for all cases. A DNN model may also include operators whose precision is defined by the precisions of their inputs, e.g. add, maxpool. 
In some circumstances, a change in operator's 
precision causes a cascading precision shift among the subsequent operators, significantly changing the overall training time.

We model the costs of converting between different precisions and between floating-point and fixed-point numbers. Casting between floating-point numbers can be modeled as a linear function of tensor size~\cite{campo}, we focus more on shaping the fixed-point case. Fixed-point quantization requires a maximize and minimize to calculate the quantization-related scaling factor. 
The process has two steps: partitioning the task into thread blocks, with each thread in a block finding the maximum and minimum of a portion of the tensor data and storing the results in a shared cache. Then, a tree-like parallel reduction is applied among all thread blocks, reducing both the number of GPU threads and data simultaneously. Thus, the cost of data collection for each step can be modeled as a linear function of the tensor size. 
Further, runtime fixed-point quantization includes calculation of the output scaling factor; if there is no fusion for the dequantization operation 
on the fixed-point execution operator, a dequantization cost should also be modeled and added
. Besides, different fixed-point quantization methods (e.g., channel-wise\cite{lee2018quantization}, layer-wise) 
vary in performance and result in different combinations of the dequantization methods. For example, a layer-wise quantized input and a channel-wise quantized weight should be dequantized with a channel-wise dequantizer, while a layer-wise quantized input and a layer-wise quantized weight should be dequantized with a layer-wise dequantizer. Fortunately, regardless of the dequantization type, it is essentially a kernel-level element-wise operation, so it can still be shaped as the linear cost with respect to the tensor size. In 
\sysname{}, we comprehensively analyze all these scenarios and employ a collection of linear models to accurately predict the casting costs across various cases, leveraging the tensor size as a parameter. 

We categorize all operators in a DNN model into two types
: 1) Precision Adjustable Operators $O_{adj}$, including common computation-intensive operators, e.g. Matmul and Conv, and operators that may numerically overflow in calculation with the low-precision number, e.g. softmax
. 
2) Precision Dependent 
Operators $O_{dep}$, whose precision is determined by the precision of the input provided by other operators
, 
e.g., add and ReLU.

\begin{algorithm}[t]
\caption{CostMapping}\label{alg:cost_mapper}
\small
\begin{algorithmic}[1]
    \STATE \textbf{Input}: Local precision DAG $\mathcal{G}_i$, target operator $o$, new precision $b_{io}$, profiled op cost $CC_{i}$, casting cost calculator $CP$, local data flow graph $DFG$
    \STATE \textbf{Output}: new precision DAG $\mathcal{G}_i'$ and local DFG $DFG'$ 
    \STATE $\mathcal{G}_i' = UpdateDAG(\mathcal{G}_i, b_{io})$ \COMMENT{Update $o$'s precision} 
    \STATE $preds, succs = \mathcal{G}_i'.pred\_succ(o)$ \COMMENT{Get neighbors of $o$} 
    \STATE $C_{i}^{fwd}=0, C_{i}^{bwd} = 0$ \\	
    \FOR{$p \in preds$} 
        \IF{$b_{ip} != b_{io}$} 
            \STATE $C_{i}^{fwd} \mathrel{+}= CP.predict(b_{ip}, b_{io}, shape_{p}^{output})$ 
        \ENDIF
    \ENDFOR
    \IF{$o \in O_{adj}$} 
        \STATE $C_{i}^{w} = CP.predict(32, b_{io}, shape_{o}^{weight})$ 
    \ELSE
        \STATE $C_{i}^{w} = 0$
    \ENDIF
    \STATE $b_{io}^{out} = output(b_{io})$ \COMMENT{Get operator output's precision} 

    \FOR{$s \in succs$} 
        \IF{$same(\mathcal{G}_i'.pred(s).bit) \& s \in O_{rel}$} 
            \STATE $CostMapping(\mathcal{G}_i, s, b_{io}^{out}, CC_{i}, CP, DFG)$ \COMMENT{Traverse} 
        \ELSE
            \STATE $UpdateFwd(s)$ \COMMENT{Lines 6-10: update forward casting cost} 
        \ENDIF
        \STATE $C_{i}^{bwd} \mathrel{+}= CP.predict(b_{is}, b_{io}, shape_{o}^{output})$
    \ENDFOR

    \STATE $C_{i}^{op} = CC_{i}[b_{io}]$ \COMMENT{Lookup pure operator execution cost} 
    \STATE $DFG'=UpdateDFG(DFG, C_{i}^{fwd}, C_{i}^{bwd}, C_{i}^{w}, C_{i}^{op})$ 
\end{algorithmic}
\end{algorithm}

\begin{figure*}[t]
  \centering
  \includegraphics[width=\linewidth]{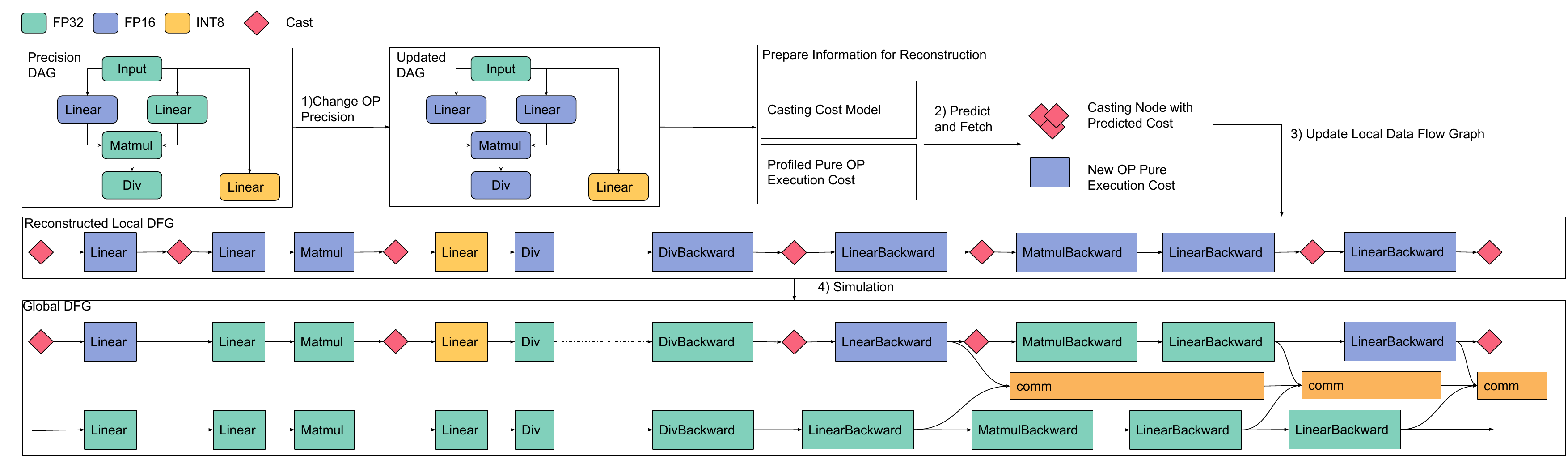}
  \vspace{-6mm}
  \caption{Workflow of Replayer. (1) The local precision DAG is updated upon a change in operator precision, and the cost mapper traverses the graph to update the precisions of dependent operators. (2) The casting costs in the new precision DAG are calculated, and the pure operator execution cost is retrieved from the profiling results. (3) the local data flow graph (DFG) is updated, and (4) the global DFG is updated accordingly, which can be used in the overall training throughput simulation.}
  \label{fig:costflow}
\vspace{-4mm}
\end{figure*}

In \sysname{}, we maintain three graphs to track the precision and execution timeline of training among different devices: Precision DAG, local DFG, and global DFG. For each GPU, \sysname{} maintains a precision DAG that keeps the training model with operators' precision and its dependencies. Each GPU also has a local DFG, which is the execution line for the operator in training, further including the backward operation and optimizers. The global DFG is composed of all the local DFGs, with communication among them.

Our cost mapper updates the precision change of an operator for a certain device to its precision DAG, computes the casting cost, and fetches new pure operator execution cost for graph update, thus producing new local DFG and global DFG. 
The new global DFG is used for training simulation. 
Alg.~\ref{alg:cost_mapper} and Fig.~\ref{fig:costflow} 
 shows the procedure of cost mapping. 
The cost mapper first updates the target operator $o$'s precision in the precision DAG $\mathcal{G}_i$ on device $i$ (line 3), and then 
records the operator's predecessors, and successors and initializes the forward cost and backward cost (lines 4-5). The casting cost is estimated by the casting cost calculator. The overall casting cost of the operator with new precision in the forward pass is calculated by summing all casting costs for input has different precisions 
$C_{i}^{fwd}$ (lines 6-10). The weight casting cost 
$C_{i}^{w}$ 
is estimated based on the operator's weight shape $shape_{o}^{weight}$ if $o \in O_{adj}$ (lines 11-13). Next, the cost mapper traverses the successor nodes of the current operator. Suppose a successor operator is a precision-dependent operator $s \in O_{rel}$ and all the input bits of the successor 
the operator is the same. In that case, breath-first-search is applied to the successors to make an iterative precision change in the precision DAG with the precision of the operator output $b_{out}$ (lines 16) \footnote{The output precision does not have to match the kernel execution precision. For example, an FP16 kernel can have an output precision of FP32 or FP16. In \sysname{}, the output of INT8 is set to a floating point FP32.} 
.The predicted backward casting cost $C_{i}^{bwd}$ is also computed (lines 17-24). Finally, the pure operator cost is fetched from the profile result (lines 25), together with new casting costs are updated to the local $DFG'$. 



\vspace{1mm}
\noindent \textbf{Simulator.}
After updating all devices with the final precision, the simulator in the Replayer simulates the execution of the global DFG. This process includes updating the communication operator cost for all devices based on communication dependencies and using topological sort to predict the execution time of all local DFGs. To trace the communication node, we first construct distributed training on smaller homogeneous GPU sets to measure the dependencies and communication buffer size. For example, to measure communication cost on a hybrid 16 T4 + 16 V100 training, we perform 2-T4 and 2-V100 training separately and use their trace data for execution timelines. Subsequently, the communication cost was recalculated by considering the topology of hybrid training and aligning the start point of the first communication, denoted as $comm_{0}^{start}$. The precise estimation of the communication cost for each local DFG on device $i$ was obtained using equation~(\ref{eq:comm_cross}), where $n$ represents the n-th communication operation:
\begin{equation}
\scalemath{0.8}{
\begin{aligned}\label{eq:comm_cross}
    comm_{n}^{start} &= max(max(\{comm_{i,n}^{start}, \forall i \in \mathcal{K} \} ), comm_{i, n-1}^{end}), \forall n \in [1,N] \\
    comm_{i,n}^{end} &= comm_{n}^{start} + max(\{comm_{i, n}^{dur} \}), \forall n \in N, \forall i \in \mathcal{K}
\end{aligned}
}
\vspace{-2mm}
\end{equation}
\noindent Here $comm_{i, n}^{start/end}$, $comm_{i, n}^{dur}$ represents the duration between start/end points and 0 and duration of the n-th communication slot on device $i$. $comm_{n}^{start}$ is the synchronized communication start point across devices.
The maximum end-to-end latency among local DFGs is then taken as the final distributed training throughput. 

Training throughput estimation, $\mathcal{E}(\cdot)$, is obtained by updating the precision cost in local DFG and precision DAG, then simulating with global DFG. 
Additionally, memory consumption of training on device $i$, $\mathcal{M}_i(\cdot)$, can be obtained simultaneously by profiling and accumulating memory consumption based on operator precision in local precision DAG $\mathcal{G}_i$.

\section{Allocator}\label{sec:allocator}
Based on the Indicator and the Replayer provided by the Predictor, the Allocator of \sysname{} solves the operator precision allocation problem~(\ref{eq:obj-funct}) to obtain operator precisions to use on the inference GPUs. 
The Allocator uses a maximum heap for each inference GPU to store indicator value differences upon precision changes. Each time, it selects the operator with the largest indicator decrement on each GPU to increase precision. Then, it estimates new overall training throughput and memory consumption using Replayer and keeps new precision if it meets memory constraints and does not decrease overall throughput. 
Instead of starting from full precision (FP32) and performing precision reduction, 
we initialize operator precisions on inference GPUs to the fastest available precision, which minimizes training time while meeting memory constraint $M^{max}_i$. 
Then we ameliorate the low-precision degradation by recovering some operators to higher-precision formats.
A list of heaps $H = \{h_i = heap_{max}(\{[\Omega_{o}^{(b_{io})} - \Omega_{o}^{(ADD(b_{io}))}, o_i]\,\mid\, \forall o \in O \})\,\mid\, \forall i \in \mathcal{K}_{inf}\}$ is maintained that records the difference of the operator indicator values under the current precision $b_{io}$ and  a higher precision\footnote{ 
For example, suppose operator $o$ on inference GPU $i$ has three precision candidates, INT8, FP16, and FP32. If $b_{io}=8$, then higher precision is FP16. 
}, together with the operator as the value. The allocator repeatedly checks if operator $o_{i}$ on an inference GPU can be increased to a higher precision without violating memory and throughput constraints, as precision change can increase memory usage and decrease training speed on the local device
; 
if so, the update is stored and a new candidate is pushed to the max heap if there exists a higher precision for the operator. 
The process iterates through the candidates in set $H$ and continues until $H$ becomes empty.

 

To find the optimal initial precision setting that maximizes training throughput, we need to consider the precision of each operator in each local DFG and the casting cost between the operator and its neighbors. An exhaustive search of the entire graph is infeasible due to the high computational complexity. Luckily, many DNN models contain repeating isomorphic building subgraphs~\cite{habitat} 
which have much fewer precision-adjustable operators available compared with the entire graph. 
(e.g. BERT's attention has only 5 such operators). We categorize the model into subgraphs and assign a memory budget to each subgraph based on its compression capacity, which is estimated by applying the lowest precision to all operators in the subgraph. 
A brute-force search is then applied to find the initial precision setting that satisfies local memory constraints while maximizing training speed.

The rationale behind the precision recovery in our allocator design is twofold. Firstly, starting with the highest-performing precision provides a reliable direction for optimization. The shortage for the other case arises from the presence of casting costs, as starting from the highest precision and reducing precision may not always result in faster speed, making it challenging to determine when to stop. Secondly, in practical terms, the precision setting that achieved the highest training throughput is often closer to the optimum. This choice reduces the number of search steps required.

\section{Backend Optimization and Implementation}




A significant challenge in implementing \sysname{}'s \scenarioname{} lies in the limited support for low-precision kernels in the existing training frameworks \cite{PyTorch}.
This includes inadequate support for low-precision fixed-point kernels (such as INT8 and INT4), as well as limitations in vendor-optimized black-box kernels 
in not supporting flexible precision changes (e.g., changing the precision of the output), which loses opportunities for optimizing low-precision kernels for improved device performance
\cite{xing2022bolt}.

To fully exploit the benefits of low-precision operators on different GPUs, we design and implement LP-PyTorch, a highly \textit{templated} backend that allows kernel configuration to the underlying lower-precision kernels for different operators. 
LP-PyTorch is designed to use the underlying kernels (e.g., CUTLASS\cite{cutlass} or CuDNN's execution kernels) in a user-friendly and precision-flexible manner. We highlight our two key designs: (1) Multi-Level Abstraction. LP-PyTorch templates each kernel as a combination of hardware-specific configuration and kernel abstractions (e.g. forward and backward pass kernels, tensor precision conversion kernel) to allow maximized flexibility and control over operators's configuration. In practice, we automatically set the composable kernel configuration, such as ThreadblockShape, WarpShape, and InstructionShape, to different precisions to optimize performance on the target hardware platform (such as GPU architecture sm70, sm75, sm80, and simt). 
(2) Front-end Security Wrapper. The tensorized kernels can have strict requirements for memory access patterns and input data precisions, e.g. TensorCore has restrictions on input tensor dimensions. We wrap kernel calls with security checks and handling using a wrap function. 

\noindent Several enhancements are included to further reduce overhead and maximize the benefits of low-precision kernels.

\noindent\textbf{Minmax Optimization.}
To calculate the scaling factor for tensor-wise fixed-point quantization, we need to find the maximum and minimum (minimax) values of a tensor. The collection process for large input shapes was observed to suffer from suboptimal GPU utilization. To address this, we developed a GPU kernel to optimize the process. We partition the process into two steps. In the first step, we collected row-wise statistics by evenly partitioning the rows (second-to-last) using a constant number of threads per block. The statistics were obtained through a warp-level primitive. Subsequently, we launched another smaller kernel to the collected row-wise results to obtain the absolute tensor-wise scalar value.

\vspace{1mm}
\noindent\textbf{Dequantization Fusion.}
In backpropagation of the low-precision kernels, we output the gradient of weight in FP32, while the gradient of activation maintains FP16 for speed up
; also, the fixed-point calculation is done in INT32 and requires additional dequantization before feeding the results into the succeeding operator
. To save the dequantization cost, we further fuse the dequantization process into the operator kernel in the epilogue level
, i.e., before copying the accumulator result into the shared memory. The lowest computation primitives of CUTLASS are done by tile iterators, \sysname{} specifies a partial iterator method 
from INT32$\rightarrow$FP32, and passes the quantization scaling factor. 




\section{Evaluation}\label{sec:eval}
\textbf{Testbed.}
We evaluate \sysname{} on real-world testbeds. (1) {\em ClusterA}: a heterogeneous cluster consists of two training servers and two inference servers. Each training server is equipped with eight Nvidia Tesla V100 GPUs with 32GB of memory and 300GB/s interconnect bandwidth. Each inference server is equipped with eight Nvidia T4 GPUs with 16GB of memory and 32GB/s interconnect bandwidth.
(2) {\em ClusterB}: a memory-constrained version of cluster A, where T4 GPUs' available memory is limited to a ratio, we set it as 30\% by default
, to emulate hybrid training scenarios in a real production system. We use all-reduce for parameter synchronization among GPUs in distributed model training. The software environment includes PyTorch-1.10.0, torchvision-0.11.0~\cite{PyTorch} for the convolution-based task, Hugging Face Transformers 4.22.0~\cite{wolf-etal-2020-transformers} for the transformer-based task the and CUDA-11.3. 

\noindent\textbf{Benchmarks.}
We mainly evaluate from-scratch training performance for convolution-based models VGG\cite{vgg16} and ResNet \cite{resnet} for image classification on ImageNet\cite{imagenet}; To show the fidelity of the predictor, we also involve transformer-based finetune task with models BERT\cite{kenton2019bert} on SQuAD\cite{rajpurkar2016squad} for question answering and RoBERTa \cite{liu2019roberta} on SWAG\cite{zellers-etal-2018-swag} for multiple-choice benchmarks
. We choose operator precisions among representative $INT8, FP16$, and $FP32$. 
Since a bit-width smaller than 16 only supports channels last (NHWC) memory format
, for a fair comparison, all convolution-based models are trained under the channels last.

\noindent\textbf{Training Configurations.} We trained VGG and ResNet models using a local batch size of 128 and a test batch size of 32, along with the SGD optimizer. The learning rate ($lr$) was set to $4.096$ for ResNet and $0.4$ for VGG. Both models underwent training for 120 epochs. For the fine-tuning of RoBERTa, we utilized a local batch size of 16 and a test batch size of 16, employing the Adam optimizer with a learning rate ($lr$) of $7.5e-5$. The fine-tuning process lasted for 6 epochs. Similarly, BERT was fine-tuned using a local batch size of 12 and a test batch size of 12, also with the Adam optimizer. The learning rate ($lr$) used was $1.2e-4$, and the training was carried out for 5 epochs.


\noindent\textbf{Baselines.} 
We compare \sysname{}'s performance with existing schemes in various aspects: (i) The end-to-end system performance (throughput) and final accuracy 
with dynamic batch sizing (DBS)~\cite{Chen2020SemidynamicLB} and uniform precision (UP), i.e., use a uniform precision for all operators in inference GPU, continue lowering precision until the memory requirement is met; We also compared with an ORACLE accuracy obtained through non-quantized (FP32) training. 
(ii) Indicator's effect with random and Hessian~\cite{hawq-v3}. 


\noindent\textbf{Metrics.} For model accuracy evaluation, we use top-1 accuracy and f1-score for classification and fine-tuning tasks, respectively, and refer to both as accuracy in the results. We evaluate final model accuracy and single-iteration training throughput following study~\cite{habitat}, 
as all our experiments are conducted under the same basic training configurations (such as the total number of epochs, 
and the learning rate scheduler).

\subsection{Performance of the Predictor}

\subsubsection{Indicator Effectiveness}\label{sec:IndicatorEff}

\begin{table}[t]
\vspace{1mm}
\centering
\resizebox{\linewidth}{!}{
\begin{tabular}{|c|cc|cc|}
\hline
\multirow{2}{*}{model} & \multicolumn{2}{c|}{ClusterA} & \multicolumn{2}{c|}{ClusterB} \\
& Method & Final Accuracy & Method & Final Accuracy \\
\hline
\multirow{2}{*}{ResNet50} & \sysname{} & \textbf{76.77(+0.24)} $\pm$ 0.43\% & \sysname{} & \textbf{76.67(+0.67)} $\pm$ 0.59\% \\
& Random & $76.53 \pm 0.53$\% & Hess & $76.00 \pm 0.43$\% \\
\hline
\multirow{2}{*}{VGG16BN} & \sysname{} & \textbf{74.77(+0.62)} $\pm$ 0.12\% & \sysname{} & \textbf{74.27(+0.91)} $\pm$ 0.06\% \\
& Random & $74.12 \pm 0.88$\% & Hess & $73.36 \pm 0.63$\% \\
\hline
\multirow{2}{*}{BERT} & \sysname{} & \textbf{87.41(+0.02)} $\pm$ 0.05\% & \sysname{} & \textbf{87.44(+0.10)} $\pm$ 0.20\% \\
& Random & $87.39 \pm 0.19$\% & Hess & $87.34 \pm 0.11$\% \\
\hline
\multirow{2}{*}{RoBERTa} & \sysname{} & $83.59 \pm 0.11$\% & \sysname{} & \textbf{82.94(+0.23)} $\pm$ 0.12\% \\
& Random & \textbf{83.61(+0.02)} $\pm$ 0.15\% & Hess & $82.71 \pm 0.31$\% \\
\hline
\end{tabular}
}
\vspace{1mm}
\caption{Indicator Performance. The best accuracy in each set of experiments is marked in bold.}
\label{tb:ind_res}
\vspace{-6mm}
\end{table}

We compare our indicator with the state-of-the-art Hessian indicator (HESS) method \cite{hawq-v3} for selecting operators in adaptive fixed-point quantization. HESS computes the block-wise Hessian for each layer and calculates the top eigenvalue, which is then divided by the parameter size and times the introduced error of the quantization. For floating-point quantization, we also compare our indicator with a random scheme. In the later approach, the largest indicator is randomly generated for the lowest precision of each operator and is halved as precision increases. For operators whose fixed-point indicator has been provided by HESS, the floating-point indicator is also halved but take it as a base. To ensure the fairness and clarity of our fixed-point quantization results, we assign different compression ratios for each trial in cluster B. These ratios are determined to emulate a 60\% maximum compression level compared to FP32 models.



Table~\ref{tb:ind_res} shows the results of our indicator. In most cases, our indicator achieves higher final model accuracy compared to existing schemes. We attribute this improved performance on cluster B to the fact that Hessian only considers weight distribution, but does not provide a comprehensive depiction of the negative impact of low-precision kernels on training.

\subsubsection{Replay Accuracy}
\begin{table}[t]

\vspace{1mm}
\centering
\resizebox{\linewidth}{!}{%
\begin{tabular}{|c|c|c|c|}
\hline
Model & Method & Avg. Est. (ms) & Err \\
 \hline
 \multirow{3}{*}{Half-Linears}& Ground Truth & 474.83 & /\\
 &w/o cost mapper(Dpro) & 427.50 & ${8 \pm 0.3}{\%}$ \\
 & \sysname & 474.52 & \textbf{3.5 $\pm$ 0.5} ${\%}$ \textbf{(-4.5)} \\
 \hline

 \multirow{3}{*}{INT-Linears}& Ground Truth & 548.46  & / \\
 &w/o cost mapper(Dpro) & 462.73 & ${13 \pm 1.9}{\%}$ \\
 & \sysname & 537.55 & \textbf{2 $\pm$ 0.1 } ${\%}$ \textbf{(-11)}\\
 \hline

 \multirow{3}{*}{Half-BertLayer1,3,5}& Ground Truth & 787.02  & /\\
 &w/o cost mapper(Dpro) & 765.55 & ${3 \pm 0.7}{\%}$  \\
 & \sysname & 781.50 & \textbf{1 $\pm$ 0.7} ${\%}$ \textbf{(-2)} \\
 \hline
\end{tabular}%
}
\vspace{1mm}
\caption{
Replay Accuracy. The best results are marked in bold.} 
\label{tb:pred_compare}
\vspace{-5mm}
\end{table}

Table \ref{tb:pred_compare} compares the predicted per-iteration training time with our predictor and DPro \cite{dpro} 
against the actual training iteration time measured. We compare the prediction results of BERT when converting all linear layers to half-precision (FP16), int8, and converting three BERT layers to half-precision. We then let Replayer estimate the latency of each of these configurations. 
Each prediction is repeated 5 times and the average prediction error is calculated. Our system's prediction error is less than 5\% in all cases, while Dpro's prediction error is much larger. We attribute it as not considering the casting costs and operator dependency 
in mixed precision training. 


        
\subsection{End-to-end Performance of QSync}\label{sec:main-exp}
\begin{figure}[t]
  \centering
  \includegraphics[width=\linewidth]{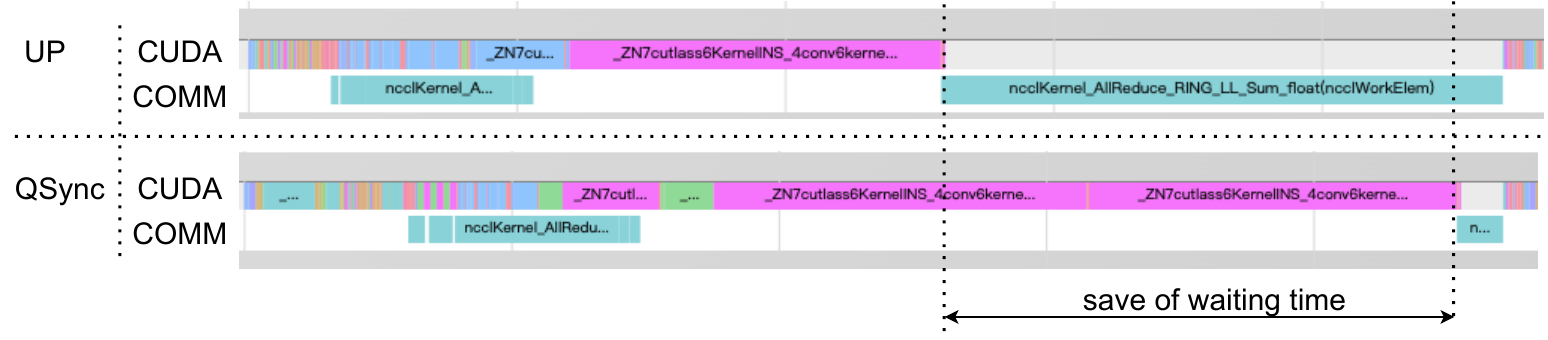}
  \caption{Training timeline of VGG16BN on ClusterA. Top: Uniform precision. Bottom: \sysname{}. \sysname{} recovers accuracy by saving the waiting time.} 
  \label{fig:sync_vgg16}
\end{figure}

\textbf{Mini Sample of \sysname{}.}
Fig.~\ref{fig:sync_vgg16} 
gives the CUDA kernel and communication timeline 
when training VGG16BN in cluster A 
using the uniform precision 
(the top execution timeline) and QSync (the bottom execution timeline). 
With our precision candidates, uniform precision accelerates computation but leads to workload mismatch, i.e. inference GPU is fully accelerated to be faster than training GPU; 
then the inference GPUs have to wait until the slow ones finish their tasks before they can start the collective communication and continue execution. 
\sysname{} recovers some of the FP16 layers to the FP32 format, greatly reducing the waiting time for communication. This improves device utilization while reducing model performance degradation caused by low-precision kernels.



\begin{table}[t]
    \centering
    \resizebox{\linewidth}{!}{%
    \begin{tabular}{|c|c|c|c|}
        \hline
        Model & Methods & Final Accuracy & Throughput (it/s) \\
        \hline
        \multirow{4}{*}{ResNet50} & ORACLE & ${76.93 \pm 0.20}{\%}$ & $\dagger$ \\
         & DBS  & ${76.13 \pm 0.05}{\%}$ & 0.40 \\
         & UP  & ${76.50 \pm 0.26}{\%}$ & 0.45 \\
         & \sysname{}  & \textbf{76.77 $\pm$ 0.43} ${\%}$ \textbf{(+0.27)} & \textbf{0.45(+0.05)} \\
        \hline
        \multirow{4}{*}{VGG16} & ORACLE & ${70.43 \pm 0.06}{\%}$ & $\dagger$ \\
         & DBS  & ${69.83 \pm 0.15}{\%}$ & 0.17 \\
         & UP  & ${69.76 \pm 0.06}{\%}$ & 0.20 \\
         & \sysname{}  & \textbf{70.33 ${\pm}$ 0.06}${\%}$ \textbf{(+0.57)} & \textbf{0.20(+0.03)} \\
        \hline
        \multirow{4}{*}{VGG16BN} & ORACLE & ${74.46 \pm 0.07}{\%}$ & $\dagger$ \\
         & DBS  & ${73.93 \pm 0.15}{\%}$ & 0.32 \\
         & UP  & ${73.80 \pm 0.10}{\%}$ & 0.38 \\
         & \sysname{}  & \textbf{74.77 $\pm$ 0.12}${\%}$ \textbf{(+0.96)} & \textbf{0.38(+0.06)} \\
        \hline
    \end{tabular}%
    }
    \vspace{1mm}
    \caption{Performance of from-scratch training in ClusterA. it/s is the iteration per second, which means how many iterations can be finished within a second. Best results are marked in bold.}
    \label{tab:clusterA_res}
    \vspace{-4mm}
\end{table}

\begin{table}[t]
    \centering
    \resizebox{\linewidth}{!}{%
    \begin{tabular}{|c|c|c|c|}
        \hline
        Model & Methods & Final Accuracy & Throughput (it/s) \\
        \hline
        \multirow{4}{*}{ResNet50} & ORACLE & ${76.93 \pm 0.20}{\%}$ & $\dagger$ \\
         & DBS  & ${76.40 \pm 0.10}{\%}$ & 0.40 \\
         & UP  & ${76.36 \pm 0.20}{\%}$ & 0.40 \\
         & \sysname{}  & \textbf{76.67 $\pm$ 0.59} ${\%}$ \textbf{(+0.33)} & \textbf{0.45(+0.05)} \\
        \hline
        \multirow{4}{*}{VGG16BN} & ORACLE & ${74.46 \pm 0.07}{\%}$ & $\dagger$ \\
        & DBS  & ${73.93 \pm 0.15}{\%}$ & 0.32 \\
         & UP  & ${73.23 \pm 0.13}{\%}$ & 0.38 \\
         & \sysname{} & \textbf{74.26 $\pm$ 0.06} ${\%}$ \textbf{(+1.03)} & \textbf{0.38(+0.06)} \\
        \hline
    \end{tabular}%
    }
    \vspace{1mm}
    \caption{Performance of from-scratch training in ClusterB. Best results are marked in bold. 
    }
    \label{tab:clusterB_res}
\vspace{-4mm}
\end{table}

\textbf{Performance of \sysname{}.}
Table \ref{tab:clusterA_res} presents a comprehensive comparison of training outcomes using \sysname{}, dynamic batch sizing, and uniform precision in the context of ClusterA. In comparison to uniform precision, the adoption of \sysname{} yields superior final accuracy while maintaining consistent throughput. Notably, for the VGG16BN model, an accuracy improvement of up to 0.96\% is observed, surpassing even that attained by single-precision. Furthermore, in terms of throughput, our system consistently achieves a gain of over 10\% when compared to dynamic batch sizing across all tasks.

In the context of ClusterB, as demonstrated in Table \ref{tab:clusterB_res}, \sysname{} consistently outperforms dynamic batch sizing and uniform precision in terms of model accuracy. Notably, it even achieves a throughput gain compared to uniform precision in the case of ResNet50. This discrepancy becomes more pronounced due to the limited availability of GPU memory, necessitating the adoption of INT8 quantization. It is worth noting that the degradation introduced by INT8 is more severe in comparison to FP16, compounded by the quantization overhead. However, \sysname{} effectively addresses this challenge by recovering unnecessary INT8 operators to their higher precision format, thereby attaining improvements in both accuracy and, remarkably, throughput.

\subsection{Performance of Transformer-Based Fine-tune Task}

\begin{table}[t]
    \centering
    \resizebox{\linewidth}{!}{%
    \begin{tabular}{|c|c|c|c|}
        \hline
        Model & Methods & Final Accuracy & Throughput (it/s) \\
        \hline
        \multirow{4}{*}{Bert} & ORACLE & ${87.49 \pm 0.08}{\%}$ & $\dagger$ \\
         & DBS & \textbf{87.52 $\pm$ 0.20}${\%}$ & 1.68 \\
         & UP  & ${87.28 \pm 0.28}{\%}$ & 1.78 \\
         & \sysname{}  & ${87.41 \pm 0.05}{\%}$ \textbf{(+0.13)} & \textbf{1.78(+0.10)} \\
        \hline
        \multirow{4}{*}{RoBERTa} & ORACLE & ${83.95 \pm 0.05}{\%}$ & $\dagger$ \\
         & DBS  & \textbf{83.73 $\pm$ 0.21}${\%}$ & 1.10 \\
         & UP  & ${83.46 \pm 0.09}{\%}$ & 1.34 \\
         & \sysname{}  & ${83.59 \pm 0.11}{\%}$ \textbf{(+0.13)} & \textbf{1.34(+0.24)} \\
        \hline
    \end{tabular}%
    }
    \vspace{1mm}
    \caption{Performance of fine-tuning tasks in ClusterA.}
    \label{tab:clusterA_finetune}
    \vspace{-4mm}
\end{table}
As depicted in Table~\ref{tab:clusterA_finetune}, the \sysname{} technique demonstrates consistent speed improvement in quantization while achieving enhanced accuracy when compared to uniform precision. However, it falls short in accuracy compared to dynamic batch sizing. We attribute this discrepancy to the inherent dissimilarities in both the structural aspects and the nature of the tasks involved. Specifically, convolution tasks employ operators that are sensitive to batch size, such as Batch Normalization (BN), whereas transformer tasks utilize Layer Normalization, which is not influenced by batch size variations. Furthermore, it is worth noting that finetuning tasks exhibit less sensitivity to batch size changes in comparison to from-scratch tasks.

\subsection{System Optimization}\label{sec:sys_opt}

\begin{figure}[t]
    \centering
    \subfigure[]{
    \includegraphics[width=0.45\linewidth]{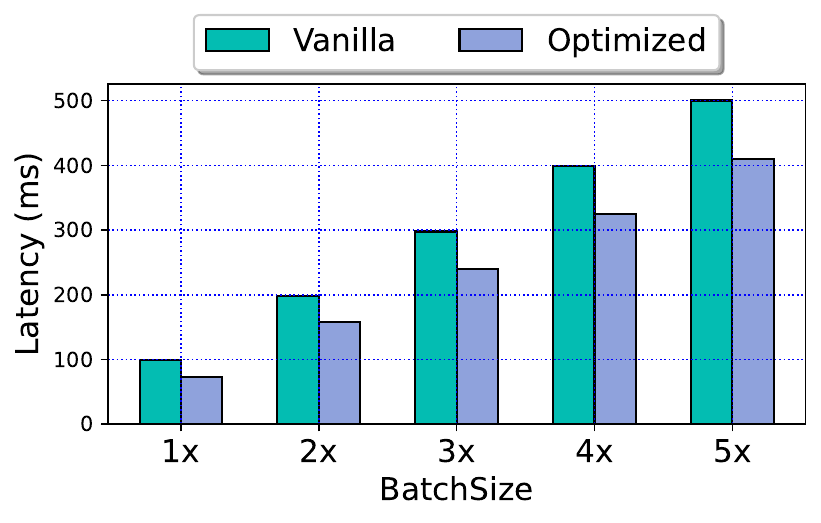}
    }
    \hspace{-3mm}
    \subfigure[]{
    \includegraphics[width=0.45\linewidth]{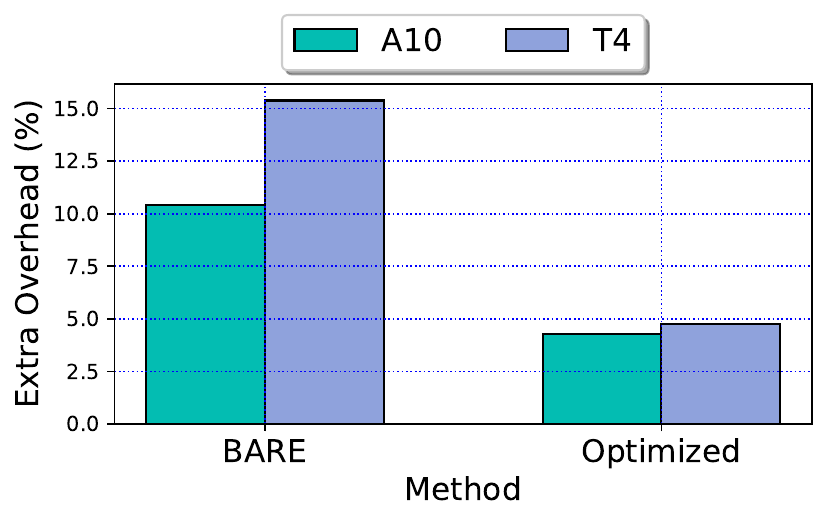}
    }
\caption{(a) Quantization Overhead Comparison for Fixed Quantization. (b) Extra Overhead Comparison for INT8 With Respect To FP16.}
\label{fig:opt_sys}
\vspace{-3mm}
\end{figure}

We evaluated to assess the effectiveness of the techniques implemented in LP-PyTorch in enhancing system efficiency. In Fig.~\ref{fig:opt_sys} (a), we quantified the quantization overhead for a tensor with a shape of $(64, 56, 56)$ and a base batch size of 64, comparing the vanilla implementation of quantization in PyTorch with our optimized approach. We performed five measurements for each method and calculated the average execution cost on the T4 GPU. The results demonstrate a significant overhead reduction of 16-20\% in the quantization process, particularly with larger batch sizes.

To further evaluate the impact of the optimization techniques we proposed in LP-PyTorch (calibration optimization and fusion), we compared the additional end-to-end overhead during the training of a ResNet50 model with a batch size of 256 on both the T4 and A10 GPUs using INT8. Fig.~\ref{fig:opt_sys} (b) illustrates the findings, with the overhead normalized against FP16 training. This experiment was conducted because full INT8 training is typically slower than FP16 due to the cost associated with casting. However, our proposed optimization methods successfully reduce this performance discrepancy from 10\% to 5\%, indicating improved efficiency for the low-precision fixed-point kernel utilization.

\subsection{Indicator Trace}

\begin{figure}[t]
    \centering
    \subfigure[Bert]{
    \includegraphics[width=0.48\linewidth]{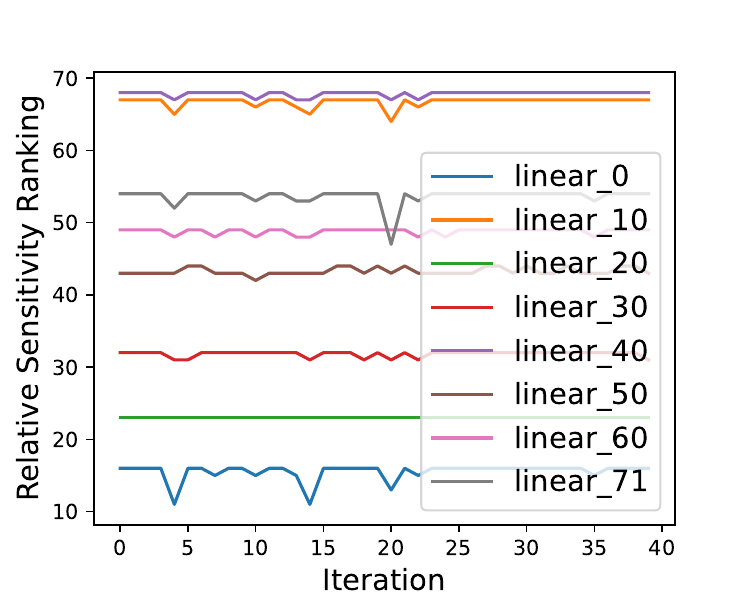}
    }
    \hspace{-4mm}
    \subfigure[ResNet50]{
    \includegraphics[width=0.48\linewidth]{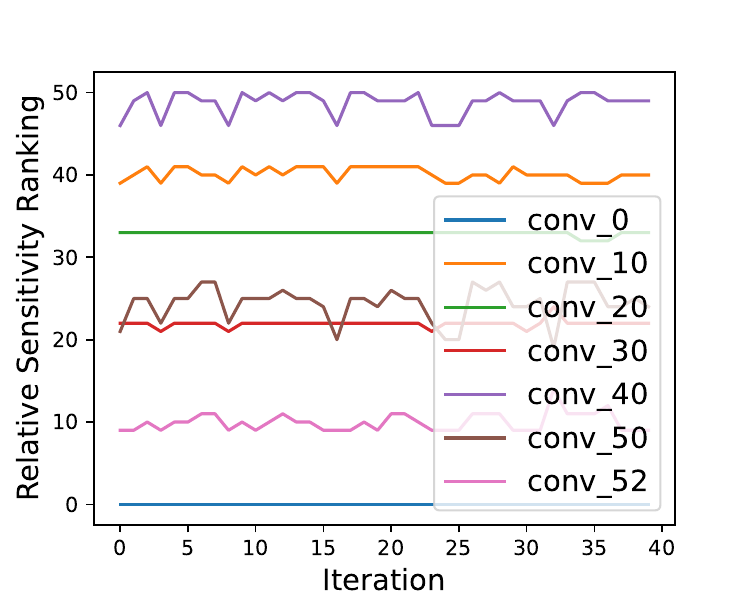}
    }
\caption{Relative Indicator Rank of (a) BERT and (b) ResNet50 for the Initial 50 Training Updates.}
\label{fig:constantupdate}
\vspace{-3mm}
\end{figure}

We conducted a comprehensive analysis of the indicator variation across multiple layers in two distinct models during the initial 50 updates of the training process. Our empirical findings demonstrate that, while fluctuations were observed between layers, the relative importance and ranking of the layers remained remarkably consistent. Notably, we observed significant disparities in layer sensitivity between the two models. In particular, the layers subsequent to the middle layers, such as the 40th linear or convolution layers, displayed significantly greater sensitivity when compared to the remaining layers.

\label{sec:indicator_trace}

\section{Discussion and Limitations}
\noindent\textbf{Efficient Profiling:} We recognize the considerable profiling overhead associated with the current implementation, which entails running a portion of the training process to trace communication nodes and indicator statistics. To address this, we suggest employing customized communication operations and alternative indicators that are less irrelevant to training progress, enabling more efficient estimation.

\noindent\textbf{\sysname{} Under Automated Mixed Precision:} AMP employs FP16/BF16 for both inference and training GPUs. We assert \sysname{} is still applicable, with the precision recovery target shifting from the inference GPU to the training GPU. We refer to this scenario as \concept{} training under the throughput-maximum case.

\noindent\textbf{System Interplay:} It is important to note that adding an inference GPU does not always lead to accelerated training. In our study, we assume that if the inclusion of an inference GPU does not enhance training speed or if the available memory is inadequate, the inference GPU will not be utilized or scheduled. Further investigation into the interplay between system components is left as future work.

\noindent\textbf{Quantization by Floor:} Another intriguing discovery is that replacing stochastic quantization with simple flooring can also restore the training quality. Investigating this further is left for future research.

\section{Conclusion}
We present \sysname{}, a \concept{} training system for \scenarioname{}. \sysname{} introduces a Predictor with an Indicator that guides operator precision selection and a Replayer that accurately simulates distributed mixed-precision training. \sysname{}'s Allocator interacts with the Predictor to decide efficient low-precision assignments to operators. 
Implemented on our optimized LP-PyTorch backend, \sysname{} provides access to a wide range of target-tuned low-precision kernels. Through empirical evaluations on various DNN models in real-world training environments, our results demonstrate that \sysname{} effectively mitigates accuracy degradation caused by low-precision operators while maintaining training throughput efficiency. 

\newpage

\bibliographystyle{IEEEtran}
\bibliography{paper}

\clearpage
\appendix

\subsection{Proof}\label{sec:appendix_proof}
\subsubsection{Proposition 1}
\begin{proof}
\sysname{} does not perform compression on any non-linear operators. Instead, it focuses on quantizing unary-input layers such as convolution and linear operators in the network. Specifically, \sysname{} does nothing with matmul ops (binary inputs). Denote the linear forward propagation function $H$. Where $H(\cdot; \mathbf{x})$ performs full-precision kernel, while $H(\cdot; \tilde{\mathbf{x}})$ performs low-precision kernel. The linear function can be written as:
\begin{equation}
    \begin{aligned}
        H^{(l)}(\mathbf{v}^{(l-1)}; \mathbf{x}^{(l)}) = \mathbf{v}^{(l-1)}\mathbf{x}^{(l)} = \mathbf{v}^{(l)}
    \end{aligned}
\end{equation}
Denote $\tilde{\mathbf{v}}$ the activation output with a low-precision kernel in the former propagation pass. We have $\tilde{\mathbf{v}}^{(0)} = \mathbf{v}^{(0)}$ thus $\mathop{\mathbb{E}}[\tilde{\mathbf{v}}^{(0)}] = \mathop{\mathbb{E}}[\mathbf{v}^{(0)}]$. Since $\mathbf{v}, \mathbf{x}$ are independent, we can easily have $\mathop{\mathbb{E}}[\tilde{\mathbf{v}}^{(l)}] = \mathop{\mathbb{E}}[\mathbf{v}^{(l-1)}]\mathop{\mathbb{E}}[\hat{\mathbf{x}}] =  \mathop{\mathbb{E}}[\mathop{\mathbb{E}}[\mathbf{v}^{(l-1)}]\mathbf{x}] = \mathop{\mathbb{E}}[\mathbf{v}^{(l)}]$. The result can be also obtained in~\cite{micikevicius2017mixed}. 
Especially, since we only consider the CE and MSE loss function and no quantization is applied to these operators, we also have the following equation for the gradient of the last input to the loss function: 
\begin{equation}
    \begin{aligned}
        \mathop{\mathbb{E}}[\nabla{\tilde{\mathbf{v}}^{(L)}}] = \lambda (\mathop{\mathbb{E}}[\tilde{\mathbf{v}}^{(L)}] - \mathop{\mathbb{E}}[y^{(L)}]) = \lambda  (\mathop{\mathbb{E}}[{\mathbf{v}}^{(L)}] - \mathop{\mathbb{E}}[y^{(L)}]) = \mathop{\mathbb{E}}[\nabla{\mathbf{v}^{(L)}}]
    \end{aligned}
\label{eq:proof_grad_last}
\end{equation}
Our gradient descent can be written as:
\begin{equation}
    \nabla  \tilde{\mathbf{x}}^{o}, \nabla \tilde{\mathbf{v}}^{o-1} = G[\nabla \tilde{\mathbf{v}}^{o}, \hat{\mathbf{C}}({\mathbf{v}}, {\mathbf{x}})]
\end{equation}

Where $\nabla \tilde{\mathbf{x}}^{o}$ is the quantization-affected gradient of the parameter at operator o, $\nabla{\tilde{\mathbf{v}}}^{o}$ is the quantization-affected gradient of activation at operator ${o}$. $G$ is the linear gradient function and $\hat{\mathbf{C}}$ context with compression. Considering equation~\ref{eq:proof_grad_last}, as $G$ is a linear function, its easy to have:
\begin{equation}
\begin{aligned}
        \mathop{\mathbb{E}}[\nabla \tilde{\mathbf{x}}^{o}, \nabla \tilde{\mathbf{v}}^{o-1}] =  \mathop{\mathbb{E}}[G[\nabla \tilde{\mathbf{v}}^{o}, \hat{\mathbf{C}}({\mathbf{v}}, {\mathbf{x}})]] &= \\ G[\nabla \mathbf{v}^{o}, \mathbf{C}(\mathbf{v}, \mathbf{x})] = \mathop{\mathbb{E}}[\nabla \mathbf{x}^{o}, \nabla \mathbf{v}^{o-1}]
\end{aligned}
\end{equation}
Which gives $\mathop{\mathbb{E}}[\nabla \tilde{\mathbf{x}}] = \mathop{\mathbb{E}}[\nabla f_s(\mathbf{x};\{b_{io}\,\mid\, o \in O \}))] = \mathop{\mathbb{E}}[\nabla f_s^{(0)}(\mathbf{x})]$. 
\end{proof}

\subsubsection{Proposition 2}


The floating-point stochastic quantization is just a rounding operation, as $\hat{x} = \bar{x} = \sround{x}$. 
\vspace{-2mm}
\begin{equation}
\scalemath{0.8}{
\sround{x} =\left\{
\begin{aligned}
&s \cdot 2^e \cdot ( 1 + \floor{m}{} + \epsilon) , &\ w.p.\ \frac{m - \floor{m}{}}{\epsilon}, \\
&s \cdot 2^e \cdot ( 1 + \floor{m}{}) , &\ w.p.\ 1 - \frac{m - \floor{m}{}}{\epsilon},
\end{aligned}
\right.
}
\vspace{-3mm}
\end{equation}\label{eq:fp_sr}

\noindent 
The mantissa $m$ is represented with $k'$ bits. $\floor{\cdot}{}$ truncates $m$ into a representation of $k$ bit, 
$\epsilon$ here is $2^{-k}$. $k = 9$ for float16. 


\begin{proof}[Variance of the Fixed-Point Quantization]

    In the context of \sysname{}, the inter-layer dataflow is conducted in floating-point format (FP16 or FP32). As a result, during the forward propagation, the execution flow of the fixed-point kernel should follow this pattern:
    \begin{equation}
    \begin{aligned}
        \hat{y}_f &= q_vq_w \sum(v_q + zp_v)(w_q + zp_w) + b_f \\ &= \sum(v_q + zp_v)q_v(w_q + zp_w)w_q + b_f \\
        &= \sum \hat{v}\hat{w} + b_f 
    \end{aligned}
    \end{equation}
    Where $q_v, q_w$ is the scaling factor, $v_q, w_q$ is the tensor in their fixed-point format. $b_f$ is the bias. Variation of $\hat{y}$ is related to the de-quantized version of fixed-point $\hat{v} \hat{w}$. In the backpropagation, the fixed-point tensor should be also de-quantized. Thus, the actual variance introduced is related to the de-quantized version of the tensor, which is related to the $q_v, q_w$.

    For fixed-point qantization, 
    we have $x - \floor{x}{} = \sigma \sim Uniform(0,1)$. This gives a $Var[\sround{\bar{\mathbf{x}}}] = \frac{D}{6}$ and $Var[\hat{\mathbf{x}}] = \frac{q_x^2D}{6}$, where $D_x$ is the dimension of the x, following the proof of EXACT \cite{liu2022exact}.
\end{proof}

\begin{proof}[Variance of Floating-point Quantization]
    By the definition of variance, we have: 
    \begin{equation}
        \begin{aligned}
             Var[\hat{\mathbf{x}}] &= \mathop{\mathbb{E}}[\hat{\mathbf{x}}^\intercal{}\hat{\mathbf{x}}] - \mathop{\mathbb{E}}[\hat{\mathbf{x}}]^\intercal{}\mathop{\mathbb{E}}[\hat{\mathbf{x}}] \\
             &= 2^{2e} \sum_{i}^D{( (1 + \floor{m}{} + \epsilon)^2 (\frac{m - \floor{m}{}}{\epsilon})} \\
             &+ {(1 + \floor{m}{})^2 (1 - \frac{m - \floor{m}{}}{\epsilon}) - (1+m)^2 )}
        \end{aligned}
    \end{equation}
    
    Take the mantissa part with new annotation $\floor{h} = 1 + \floor{m}, h = 1 + m$, and we have $h - \floor{h}{} = \varepsilon \sim Uniform(0,\epsilon) = Uniform(0,2^{-k})$. The variance can be written as:
    
    \begin{equation}
    \begin{aligned}
         Var[\hat{\mathbf{x}}] &= 2^{2e} \sum_{i}^D{(\floor{h} + \epsilon)^2(\frac{h - \floor{h}}{\epsilon}) + \floor{h} ^ 2 (1-\frac{h-\floor{h}}{\epsilon} - h^2} \\
         &=2^{2e} \sum_{i}^D{2h\floor{h} + \epsilon{h} - \floor{h} ^2 - \epsilon \floor{h} - h^2} \\
         &=2^{2e} \sum_{i}^D{\sigma \epsilon - \sigma^2}
    \end{aligned}
    \end{equation}
    
    Take expectation w.r.t. $\sigma$ on both sides, we have $Var[\hat{\mathbf{x}}] = \frac{2^{2e} \epsilon^2 D}{6}$

\end{proof}

In practice, a technique called loss-scaling may be applied to prevent gradient values from becoming too small when using mixed precision. This technique involves multiplying the loss value by a scaling factor called $S_{loss}$. However, this scaling factor is applied to all operators, we focus on analyzing the relative quantization sensitivity among operators and ignore it in our modeling of single-operator variance.

\subsubsection{Proposition 3}

\begin{proof}
This proof is adopted from ActNN \cite{chen2021actnn}. The differences compared with ActNN are 1. We introduce precision loss for both input/gradient of output and weight 2. We consider both forward pass and backward pass. Given a non-arbitrary model structure, its corresponding forward DAG, and the deepest operator with depth $L(d_L)$, we define activation forward method $H$ and the following propositions:


\begin{equation}
\begin{aligned}
    &H^{l \sim m}(\tilde{\mathbf{v}}^{(m-1)}) = \\
    &H^{(l)}(H^{(l-1)}(\cdots H^{(m)}(\tilde{\mathbf{v}}^{(m-1)}; \mathbf{\mathbf{x}}^{(m)}) \cdots, {\mathbf{x}}^{(l-1)}, \mathbf{x}^{(l)}) \\
    &H^{l \sim m}(\tilde{\mathbf{v}}^{(m-1)}; \tilde{\mathbf{x}}^{(m)}) = \\
    &H^{(l)}(H^{(l-1)}(\cdots H^{(m)}(\tilde{\mathbf{v}}^{(m-1)}; \tilde{\mathbf{x}}^{(m)}) \cdots, {\mathbf{x}}^{(l-1)}, \mathbf{x}^{(l)})
\end{aligned}
\end{equation}
\sysname{} didn't change the precision of the binary-inputs operator. This also allows us to linearly (layer-wisely) analyze the variance increment of each operator with depth $l(d_l)$. We can easily have:
\begin{equation}
    Var[H^{l \sim 1}(\tilde{\mathbf{v}}^{(0)})] = Var[H^{l \sim 1}({\mathbf{v}}^{(0)})] = Var[\mathbf{\mathbf{v}}^{(l)}]
\end{equation}
For $m > 1$, due to law of total variance $Var[x] = \mathop{\mathbb{E}}[Var[x\,\mid \,y]] + Var[\mathop{\mathbb{E}}[x\ |\ y]]$, 

\begin{equation}
\begin{aligned}
    &Var[H^{l \sim m}(\tilde{\mathbf{v}}^{(m-1)})] = H^{l \sim m}(H^{m-1}(\tilde{\mathbf{v}}^{(m-2)}; \tilde{\mathbf{x}}^{(m-1)})) \\
    &= Var[\mathop{\mathbb{E}}[H^{l \sim m}(H^{m-1}(\tilde{\mathbf{v}}^{(m-2)}; \tilde{\mathbf{x}}^{(m-1)})) | \tilde{\mathbf{v}}^{(m-2)}]] \\
    &+ \mathop{\mathbb{E}}[Var[H^{l \sim m}(H^{m-1}(\tilde{\mathbf{v}}^{(m-2)}; \tilde{\mathbf{x}}^{(m-1)})) | \tilde{\mathbf{v}}^{(m-2)}]] \\
    &= Var[H^{l \sim m-1}(\tilde{\mathbf{v}}^{(m-2)})] \\
    &+ \mathop{\mathbb{E}}[Var[H^{l \sim m}(H^{m-1}(\tilde{\mathbf{v}}^{(m-2)}; \tilde{\mathbf{x}}^{(m-1)})) | \tilde{\mathbf{v}}^{(m-2)}]] \\
    &= Var[\mathbf{\mathbf{v}}^{(l)}] + \sum_{k=1}^{l}\mathop{\mathbb{E}}[Var[H^{l \sim k}(\tilde{\mathbf{v}}^{(k-1)}; \tilde{\mathbf{x}}^{(k)}) | \tilde{\mathbf{v}}^{(k-1)}]]
\end{aligned}
\end{equation}

In the forward propagation, this gives:
\begin{equation}
\begin{aligned}
    &Var[\tilde{\mathbf{v}}^{(l)}] = Var[H^{l \sim l}(\tilde{\mathbf{v}}^{(l-1)})] \\
    &= Var[\mathbf{\mathbf{v}}^{(l)}] + \sum_{k=1}^{l}\mathop{\mathbb{E}}[Var[H^{l \sim k}(\tilde{\mathbf{v}}^{(k-1)}; \tilde{\mathbf{x}}^{(k)}) | \tilde{\mathbf{v}}^{(k-1)}]]
\end{aligned}
\end{equation}
Similarly, we define the gradient method of activation and the parameter to be $\nabla{\mathbf{v}^{(l-1)}}, \nabla{\mathbf{x}^{(l)}} = G[\nabla{\mathbf{v}^{(l)}}, \mathbf{C}(\mathbf{v}^{(l-1)}, \mathbf{x}^{(l)})]$
We have:
\begin{equation}
\begin{aligned}
    &G_{\mathbf{x}}^{l \sim m}(\nabla \mathbf{v}^{(m)}) = \\
    &G_{\mathbf{x}}^{(l)}(G_{\mathbf{v}}^{(l+1)}(\cdots G_{\mathbf{v}}^{(m)}(\nabla {\mathbf{v}}^{(m)}; \mathbf{C}^{(m)}) \cdots; {\mathbf{C}}^{(l+1)}); {\mathbf{C}}^{(l)}) \\
    &G_{\mathbf{x}}^{l \sim m}(\nabla \mathbf{v}^{(m)}; \hat{\mathbf{C}}^{(m)}) = \\
    &G_{\mathbf{x}}^{(l)}(G_{\mathbf{v}}^{(l+1)}(\cdots G_{\mathbf{v}}^{(m)}(\nabla {\mathbf{v}}^{(m)}; \hat{\mathbf{C}}^{(m)}) \cdots; {\mathbf{C}}^{(l+1)}); {\mathbf{C}}^{(l)})
\end{aligned}
\end{equation}
Hence, for $m < L$, we have
\begin{equation}
\begin{aligned}
    &Var[G_{\mathbf{x}}^{l \sim m}(\nabla \tilde{\mathbf{v}}^{(m)})] = Var[G_{\mathbf{x}}^{l \sim m+1}(\nabla \tilde{\mathbf{v}}^{(m+1)})] \\
    &+ \mathop{\mathbb{E}}[Var[G_{\mathbf{x}}^{l \sim m}(G_{\mathbf{v}}^{m+1}(\nabla \tilde{\mathbf{v}}^{(m+1)}; \hat{\mathbf{C}}^{(m+1)})] | \nabla \tilde{\mathbf{v}}^{(m+1)}] \\
    &= Var[G_{\mathbf{x}}^{l \sim L}(\nabla \tilde{\mathbf{v}}^{(L)})] \\ 
    &+ \sum_{k=m+1}^{L}\mathop{\mathbb{E}}[Var[G_{\mathbf{x}}^{k\sim{L}}((\nabla \tilde{\mathbf{v}}^{(k)}; \hat{\mathbf{C}}^{k})]   | \nabla \tilde{\mathbf{v}}^{(k)})]
\end{aligned}
\end{equation}
And,
\begin{equation}
\begin{aligned}
    Var[G_{\mathbf{x}}^{l \sim L}(\nabla \tilde{\mathbf{v}}^{(L)}; \tilde{\mathbf{v}}^{(l)})] &= Var[\tilde{\mathbf{v}}^{(l)}]
\end{aligned}
\end{equation}
This gives:
\begin{equation}
\begin{aligned}
    &Var[\nabla{}\tilde{\mathbf{x}}^{(l)}] = \\
    &Var[\nabla{}\tilde{\mathbf{v}}^{(l)}] + \sum_{k=m+1}^{L}\mathop{\mathbb{E}}[Var[G_{\mathbf{x}}^{l \sim k}(\nabla \tilde{\mathbf{v}}^{(k)}; \hat{\mathbf{C}}^{k}) | \nabla \tilde{\mathbf{v}}^{(k)}]]
\end{aligned}
\end{equation}

\subsubsection{Theorem 2}
\begin{proof}
    Given the gradient function of CE and MSE, the gradient variance of the last layer can be expressed with  $Var[\nabla{{\mathbf{v}}}^{(L)}]= \gamma^2 Var[{\mathbf{v}}^{(l)}]$.
\end{proof}
This gives the final variance increment indicator for the layer, which is:
\begin{equation}
\begin{aligned}
    &Var[\nabla{}\tilde{\mathbf{x}}^{(l)}] = \gamma^2 Var[{\mathbf{v}}^{(l)}] + \gamma^2 \sum_{k=1}^{l}\mathop{\mathbb{E}}[Var[H^{l \sim k}(\tilde{\mathbf{v}}^{(k-1)}; \tilde{\mathbf{x}}^{(k)}) | \hat{\mathbf{v}}^{(k-1)}]] \\
    &+ \sum_{k=m+1}^{L}\mathop{\mathbb{E}}[Var[G_{\mathbf{x}}^{l \sim k}(\nabla \tilde{\mathbf{v}}^{(k)}; \hat{\mathbf{C}}^{k}) | \nabla \tilde{\mathbf{v}}^{(k)}]] \\
    &= \gamma^2 Var[{\mathbf{v}}^{(l)}] + \gamma^2\sum_{k=1}^{l}\tilde{\sigma}_{fp}^{(k)} + \sum_{k=l+1}^{L}\tilde{\sigma}_{bp}^{(k)}
\end{aligned}
\end{equation}
The two terms $\hat{\sigma}^{(l)}$ signify the variance increment resulting from the utilization of a low-precision kernel. It is important to note that we analyze the bounds for these terms, considering a relaxed assumption of independence between a layer's input and the output of its ancestor/predecessor during both forward and backward propagation. By leveraging the relationship $Var[XY] = E[Y]^2 Var[X] + E[X]^2Var[Y]$ and Proposition 2, we derive the variance increment for the core operators within \sysname{}.
\vspace{-1mm}
\begin{equation}
\scalemath{0.8}{
\hat{\sigma}_{fp} = \left\{
\begin{aligned}
&\frac{1}{6} (\|{\mathbf{x}}\|^2 q_{\tilde{{\mathbf{v}}}}^2D_{{\mathbf{v}}} + \|\hat{{\mathbf{v}}}\|^2 q_{{\mathbf{x}}}^2D_{{\mathbf{x}}}), &\ \mbox{\small fixed-point quantization}, \\
& \frac{1}{6}\epsilon^2(\|\mathbf{x}\|^22^{2e_{\mathbf{v}}}D_{\mathbf{v}} + \|\hat{{\mathbf{v}}}\|^22^{2e_{\mathbf{x}}}D_{\mathbf{x}}), &\ \mbox{\small floating-point quantization}
\end{aligned}
\right.
}
\end{equation}

\vspace{-7mm}

\begin{equation}
\scalemath{0.8}{
\hat{\sigma}_{bp} = \left\{
\begin{aligned}
&\frac{1}{6}(\|\nabla{{\mathbf{v}}}\|^2q_{\tilde{{\mathbf{v}}}}^2D_{{\mathbf{v}}} + \|\hat{{\mathbf{v}}}\|^22^{2e_\nabla{{\mathbf{v}}}}\epsilon^2D_\nabla{{\mathbf{v}}}), &\ \mbox{\small fixed-point quantization}, \\
& \frac{1}{6}\epsilon^2(\|\nabla{\hat{{\mathbf{v}}}}\|^22^{2e_{\mathbf{v}}}D_{{\mathbf{v}}} + \|\hat{{\mathbf{v}}}\|^22^{2e_\nabla{{\mathbf{v}}}}D_\nabla{{\mathbf{v}}}), &\ \mbox{\small floating-point quantization}.
\end{aligned}
\right.}
\end{equation}

We denote the variance increment for operator with depth $l$ as $\Omega_{o}^{(b_o)}$, which is:
\begin{equation}
    \begin{aligned}
        \Omega_{o}^{(b_o)} &= \gamma^2 \sum_{k=1}^{l} \hat{\sigma}_{fp}^{(k)} + \sum_{k=l+1}^{L} \hat{\sigma}_{bp}^{(k)}  \\
        &= \gamma^2 d_o\hat{\sigma}_{fp}^{(o)} + (d_L-d_o) \hat{\sigma}_{bp}^{(o)}
    \end{aligned}
\end{equation}
\end{proof}

\label{sec:appendix_simu}

\end{document}